\documentclass[lettersize,journal]{IEEEtran}
\usepackage{amsmath,amsfonts}
\usepackage{algorithmic}
\usepackage{algorithm}
\usepackage{array}
\usepackage[caption=false,font=normalsize,labelfont=sf,textfont=sf]{subfig}
\usepackage{textcomp}
\usepackage{stfloats}
\usepackage{url}
\usepackage{verbatim}
\usepackage{graphicx}
\usepackage{cite}
\usepackage[super,sort&compress,comma]{natbib}

\usepackage{fancyhdr}		

\usepackage{graphicx}

\usepackage{soul}

\makeatletter \renewcommand\@biblabel[1]{$^{#1}$} \makeatother
\renewcommand{\thesection}{{\sf \Roman{section}}.}
\renewcommand{\thesubsection}{\thesection{\sf \Alph{subsection}}.}
\renewcommand{\thesubsubsection}{\thesubsection{\sf \arabic{subsubsection}}.}

\usepackage[mathlines]{lineno}

\usepackage{hyperref}
\hypersetup{ colorlinks,
    citecolor=blue,
    filecolor=blue,
    linkcolor=blue,
    urlcolor=blue
}

\usepackage{xcolor}

\definecolor{gray}{rgb}{0.6,0.6,0.6}
\definecolor{red}{rgb}{0.85,0,0}
\definecolor{green}{rgb}{0,0.85,0}
\definecolor{blue}{rgb}{0,0,0.85}
\definecolor{beige}{rgb}{0.92,0.87,0.78}
\usepackage[all]{hypcap}    

\newcommand{\be}{\begin{equation}} 
\newcommand{\ee}{\end{equation}}   

\def \bfv{{\mathbf v}}

\usepackage{amsmath}
\usepackage{amssymb}
\usepackage{mathrsfs}

\def \bbR{{\mathbb{R}}}

\usepackage{xcolor}

\begin{document}

\title{Determination of the Human-Robot Interaction Workspace for Robot Assisted Head Stabilization Control}

\author{
Xinmin Liu, PhD$^1$,
Alexander L. Van Slyke, PhD$^2$,
Destinee Holloman$^3$,
Ahmad Sakaamini, PhD$^4$,
Rodney D. Wiersma, PhD$^1$\\
$^1$ Department of Radiation Oncology, University of California at Los Angeles, CA\\
$^2$ Department of Radiation Oncology, Yale New Haven Hospital, Trumbull, CT\\
$^3$ University of Colorado, Anschutz Medical Campus,
Aurora, CO\\
$^4$ Department of Radiation Oncology, University of Pennsylvania, Philadelphia, PA
}



\maketitle

\begin{abstract}
\textcolor{black}{
Accurate head immobilization is critical for medical procedures such as neuroimaging and intracranial radiation therapy, where even minor motion can compromise precision. Traditional methods, such as rigid stereotactic frames and thermoplastic face masks, are invasive, uncomfortable for patients, and require a time-consuming setup. To overcome these limitations, a novel approach employs a 6-degree-of-freedom (6DoF) robotic head holder that is integrated with real-time optical motion tracking, enabling dynamic compensation for head movements without physical restraint. However, implementing this system requires a detailed biomechanical understanding of skull-cervical spine interactions to define safe and effective motion limits. In this study, we conduct a human volunteer investigation using a 6DoF Stewart platform robot to characterize biokinematic workspace of the head-neck system and establish foundational data for
human-robot interaction in precision head stabilization.
}

\end{abstract}

\begin{IEEEkeywords}
stereotactic surgery, medical robotics, neck kinematics, robot human interaction
\end{IEEEkeywords}

\section{Introduction}
There has been significant interest in achieving high-precision positioning of the head in medical applications~\cite{xue2017high,dellios2021evaluation,vergalasova2019multi}. In diagnostic brain imaging, head motion can lead to movement-related artifacts such as blurring and ghosting, which can hinder clinical diagnosis~\cite{liu2018effect}. This is especially true for imaging modalities such as functional magnetic resonance imaging (MRI) and positron emission tomography (PET), where the collection of data required for 3D image reconstruction requires patients to lie still for long periods of time~\cite{zaitsev2015motion,gurney2018mri}. This issue is especially challenging when scanning children or adults who exhibit involuntary motion~\cite{copeland2021infant,woods2010training,li2023pet}. 
Prospective motion correction (PMC) is a popular technique that updates the image acquisition in real-time to keep the field of view constant relative to the moving object in brain MRI. However, PMC has some limitations. It cannot correct higher-order effects such as changing magnetic field inhomogeneities and gradient imperfections. When using multiple receiver coil arrays for better signal quality, motion can change the coil sensitivities relative to the moving object. Motion during image acquisition can cause blurring and spatial distortion because data is collected from different positions within the nonlinear gradient fields. Even with accurate real-time pose correction, motion can still cause magnetic field changes due to inhomogeneities and object-induced field changes. These local magnetic field changes are often the main source of remaining artifacts~\cite{godenschweger2016motion,zahneisen2016reverse,williams2023ultra,banerjee2013parallel,wang2023diffusion}. \textcolor{black}{Therefore,} it is desirable to use a robotic platform to stabilize the head position.

Although computed tomography (CT) can typically acquire images faster than MRI or PET, in certain cases, head motion artifacts have been reported~\cite{spin2016patient,stutzki2015indications}. In therapeutic applications, accurate head positioning is essential for procedures involving the treatment of the brain using conventional surgery or radiation therapy~\cite{krist2022management,gaspar2010role,graber2019congress}. Stereotactic radiosurgery (SRS) is a treatment modality employed for managing brain disorders by utilizing highly precise radiation dose placement. SRS has been used to treat benign or metastatic brain tumors~\cite{gaspar1997recursive}, arteriovenous malformations (AVMs)~\cite{jalali2009micromultileaf,betti1989stereotactic}, trigeminal neuralgia~\cite{zahra2009stereotactic,chen2004treatment}, and acoustic neuromas~\cite{litre2013fractionated}. For SRS, accurate head positioning is needed to avoid damaging critical neural and vascular structures within the brain.

\begin{figure}[ht!]
	\centerline{
		\includegraphics[height=36mm]{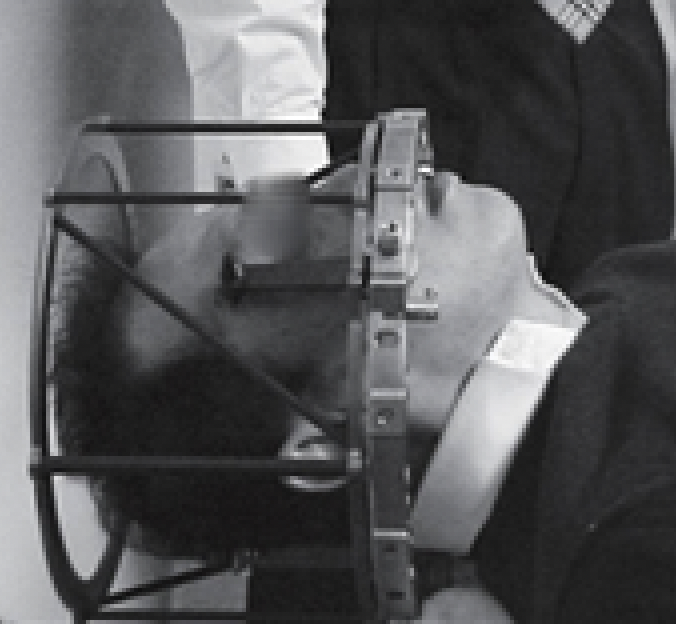}
		\includegraphics[height=36mm]{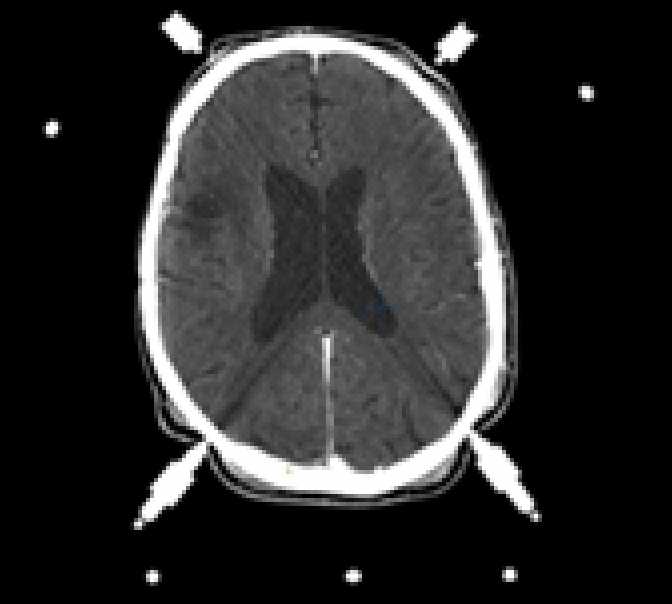}
	}
	\vspace{3mm}
	\centerline{
		\includegraphics[height=23.2mm]{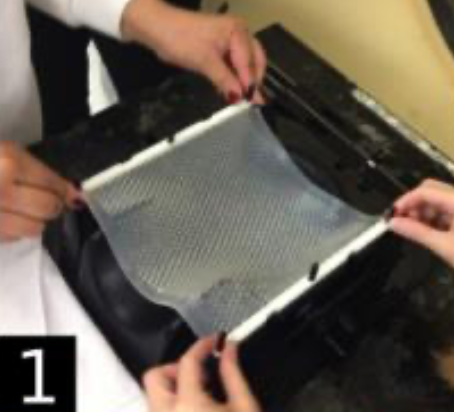}
		\includegraphics[height=23.2mm]{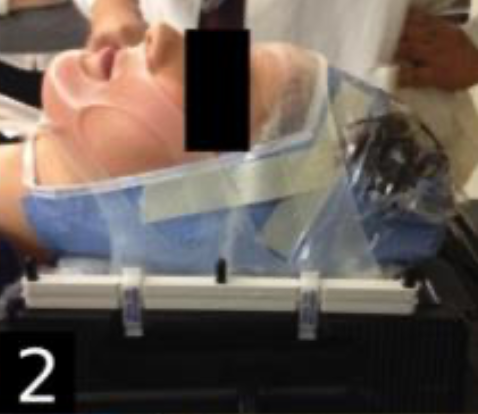}
		\includegraphics[height=23.2mm]{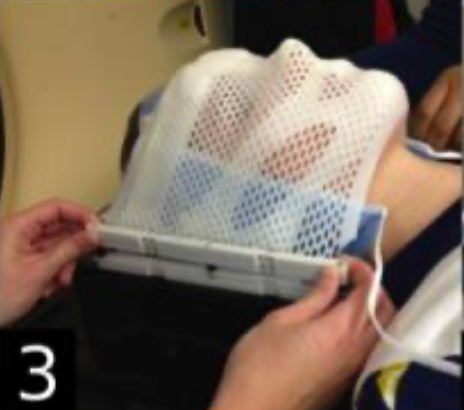}
	}
	\caption{\footnotesize 
		SRS immobilization.
		Top row: Traditional frame-based SRS setup where a metal ring is fixed to the patient's skull using four screws \cite{chelvarajah2004cranial}.
		Bottom row: Frameless SRS immobilization. A triple mask configuration to minimize head movement: (1) bottom mask, (2) intermediate mask with chin strap, and (3) top masks\cite{li2013migration}.
	}
	\label{immobilization}
\end{figure}

The most common approach to stabilize the head position is \textcolor{black}{the use} of stereotactic frames. In this approach, a metal ring is attached to the patient's skull using screws, and then the frame is bolted to the treatment table, see top row in Fig.~\ref{immobilization}~\cite{chelvarajah2004cranial,murphy2009intrafraction}. The invasive nature and discomfort associated with the frame have been identified as \textcolor{black}{a} central cause of poor patient compliance and poor clinical efficacy. Another approach involves thermoplastic face masks, as shown in \textcolor{black}{the} bottom row of Fig.~\ref{immobilization}. However, such masks have decreased accuracy due to patient-induced mask flex and can suffer from material changes over time, such as shrinkage~\cite{wagner2000geometrically}. 
\textcolor{black}{
Even with the mask in place, a patient's head can still drift laterally due to rotational movement around the fulcrum at the back of the skull. In some cases, this has resulted in systematic shifts of up to 2 mm from the initial target setup position \cite{murphy2003patterns}. A recent study reported significant errors, which were attributed to improper patient setup stemming from factors such as limitations in mask manufacturing quality, loosening or shrinkage of the mask during treatment, or physical changes in the patient's head due to facial swelling or weight loss\cite{keeling2016quantitative}. Moreover, many patients find the thermoplastic mask invasive and psychologically confining, as it severely restricts natural head motion. These factors can contribute to unsafe conditions, potentially leading to treatment interruptions for patient realignment to ensure proper target coverage\cite{murphy2009intrafraction}.
The clinical significance of localization inaccuracies in frameless stereotactic radiosurgery (SRS) remains a subject of debate. For critical applications—such as treating acoustic neuromas, pituitary adenomas, trigeminal neuralgia, or lesions adjacent to vital structures—frameless SRS methods have not yet achieved the same level of confidence as frame-based approaches. To mitigate this concern, image-guided radiotherapy (IGRT) is commonly used to monitor patient positioning, complemented by 6D robotic couch adjustments to reduce setup errors in frameless SRS\cite{takakura2009geometric, he2023frameless, almeida2020positioning}. However, IGRT is typically performed intermittently and does not compensate for real-time patient motion deviations during radiation delivery.
}
In general, the accuracies of masks are not suitable for deep tumors located near critical structures, such as \textcolor{black}{in} trigeminal neuralgia cases.

\begin{figure}[ht!]
\vspace{-5pt}
\centerline{
\includegraphics[width=80mm]{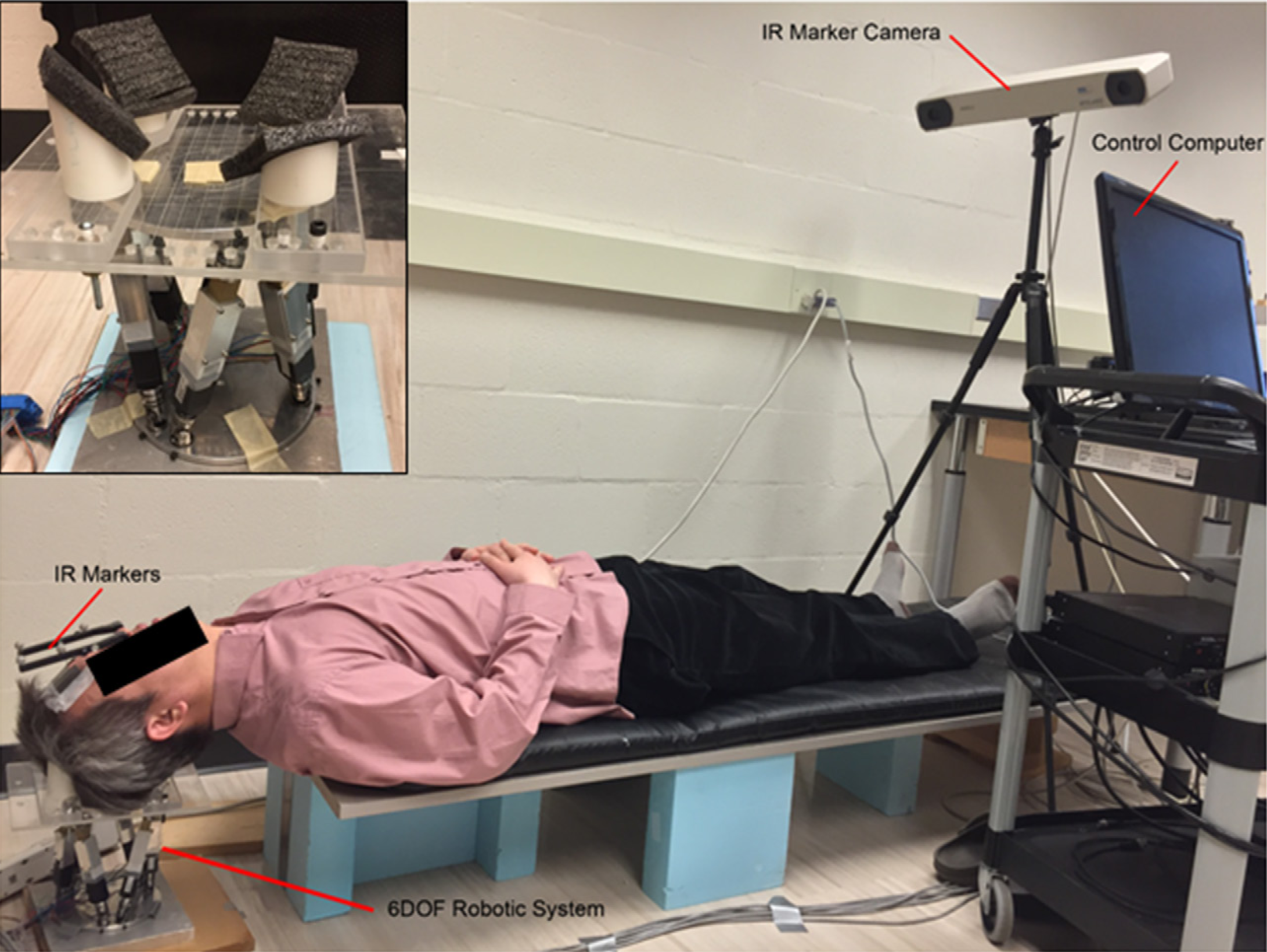}
}
\caption{\footnotesize
Robotic SRS prototype head motion stabilization for volunteers using IR marker tracking head motions. Camera inputs are automatically transformed and 6DoF errors are determined. A cancellation trajectory is calculated, and motion to accomplish cranial stabilization is transmitted to the robot's linear actuators.
}
\label{hexapodSRS}
\end{figure}

One recent method that allows highly accurate human head positioning without the use of head frames or thermoplastic masks is through the use of a 6-degree-of-freedom (6DoF) robotic head holder \cite{belcher2017towards,wiersma2019conceptual}. By performing real-time 6DoF patient head motion tracking through a suitable camera system that is used as an input into \textcolor{black}{a} control system\textcolor{black}{,} it was \textcolor{black}{demonstrated} that a Stewart platform (hexapod) type of robot can be used to continuously correct the head target position both translationally and rotationally during treatment (Fig.~\ref{hexapodSRS}). \textcolor{black}{The} 
system efficacy was evaluated on healthy volunteers and indicated that head motion remained under a 0.5mm/0.5deg threshold $>$ 99\% of the time, compared to less than 10\% of the time without the use of robotic stabilization demonstrating that the robotic SRS has potential to match or exceed the accuracies of frame based procedures. 

Currently\textcolor{black}{,} robotic head position stabilization systems do not consider the biokinematics of the head and neck, but rather treat the head as a rigid body object that can freely move in space. This control method can therefore lead to robot head correction trajectories that do not follow natural head-neck motion and can be uncomfortable for patients. 
\textcolor{black}{The} use of an accurate biomechanical model of the head and neck could allow for improved patient comfort and \textcolor{black}{more effective} robotic head motion correction. However, such kinematic modeling of the human head and neck is a complex subject that requires a multidisciplinary approach, combining knowledge of anatomy, biomechanics\textcolor{black}{,} and engineering. The study \cite{salem2013vivo} aimed to examine the three-dimensional kinematics of the cervical spine during maximal passive head rotation, with a focus on coupled motion, using low-dose CT scans of twenty healthy volunteers in both neutral and maximum axial rotation positions. The study employed \textcolor{black}{the} finite helical-axis method to describe the 3D motion between the discrete positions of each separated vertebra. In another work\textcolor{black}{,} \textcolor{black}{investigating} the spine’s role during walking\textcolor{black}{,} a 3-dimensional multi-segment kinematic spine model \cite{konz2006kinematic} was developed for noninvasive analysis of spinal motion during walking. Validation of the kinematic model was carried out by comparing the calculated angles to the known angles from a mechanical model. \textcolor{black}{The study} involved collecting data on the motion of the spine from volunteers walking at their self-selected speeds. Kinematic modeling can also be useful in the development of devices aimed at addressing issues in the head and neck region. A novel study \cite{zhang2019robotic} was reported where a dynamic neck brace was utilized to analyze the head movements of patients with amyotrophic lateral sclerosis (ALS) while also recording the surface electromyography (EMG) of their neck muscles. \textcolor{black}{The} ALS participants \textcolor{black}{were} instructed to perform three head-neck movements in a single plane: flexion-extension in the sagittal plane, lateral bending in the coronal plane\textcolor{black}{,} and axial rotation in the transverse plane. These movements were repeated five times at self-selected speeds. There are also many studies on occupational neck disorders as a result of \textcolor{black}{the workforce shifting} from industry to sedentary office work\textcolor{black}{,} and \textcolor{black}{on} the mechanical behavior of the cervical vertebra under normal conditions and under conditions that result in minor or major injuries \cite{alizadeh2020electromyography,bogduk2000biomechanics,suderman2012neck}.

In this work, we present results from a healthy volunteer study investigating head and neck reaction to a 6DoF robot that is aimed \textcolor{black}{for potential use} in SRS radiation therapy. Optical camera monitoring was used to track the real-time response of the human head to various translational and rotational input trajectories induced by a compact 6DoF robot located behind the head. Analysis of motion data allowed for determination of general kinematic parameters as needed for the development of a head-and-neck biomechanical model. Knowledge of the head's dynamic motion constraints \textcolor{black}{is an} important consideration\textcolor{black}{s} for the interaction between humans and robots\textcolor{black}{,} not only in the medical space but potentially \textcolor{black}{in} other research \textcolor{black}{areas} such as the development of \textcolor{black}{head motion control through robotic exoskeletons}.

\section{Methods and Materials}
\label{methods}

\subsection{Robotic Head Motion Control System}
\label{roboticSRS}

The robot utilizes a parallel kinematic approach and is equipped with six linear actuators to enable full 6DoF head control, including \textcolor{black}{translations in }x, y, z\textcolor{black}{,} and \textcolor{black}{rotations in }pitch, yaw, \textcolor{black}{and} roll. The six extendable legs are positioned in a precise geometrical configuration, providing several advantages over serial robotic systems. 
\textcolor{black}{The motion system features a platform with a radius of 75 mm and a stationary height of 183 mm, and provides a motion workspace of $\pm$15 mm along the XYZ axes and $\pm$15\textcolor{black}{$^\circ$} in pitch, roll, and yaw \cite{belcher2014development}.}
These advantages include the distribution of weight load across all legs, resulting in high compactness, and the division of overall positional error among all linear actuators. 

\begin{figure*}[!t]
	\centering
	\includegraphics[height=60mm]{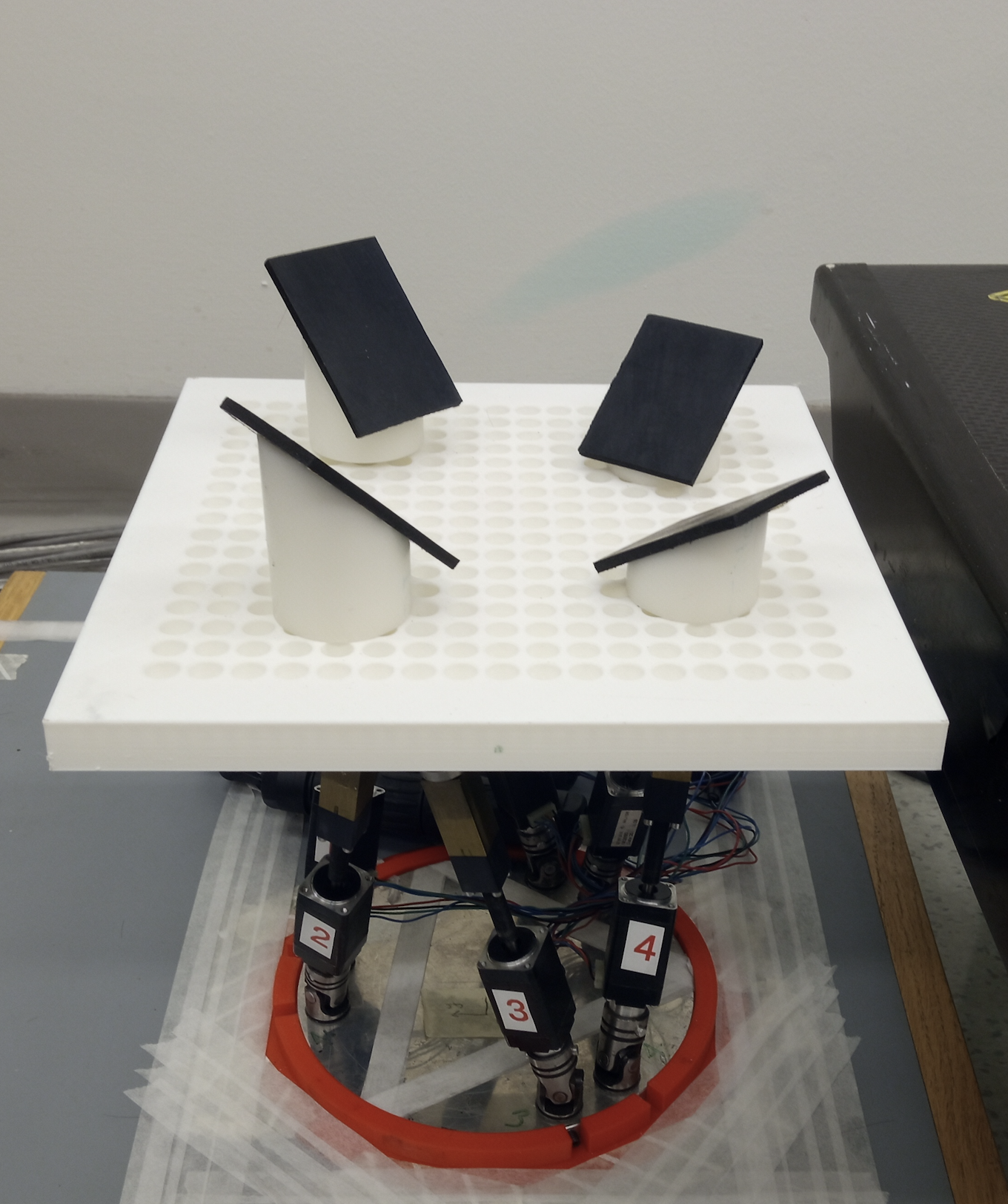}
	\includegraphics[height=60mm]{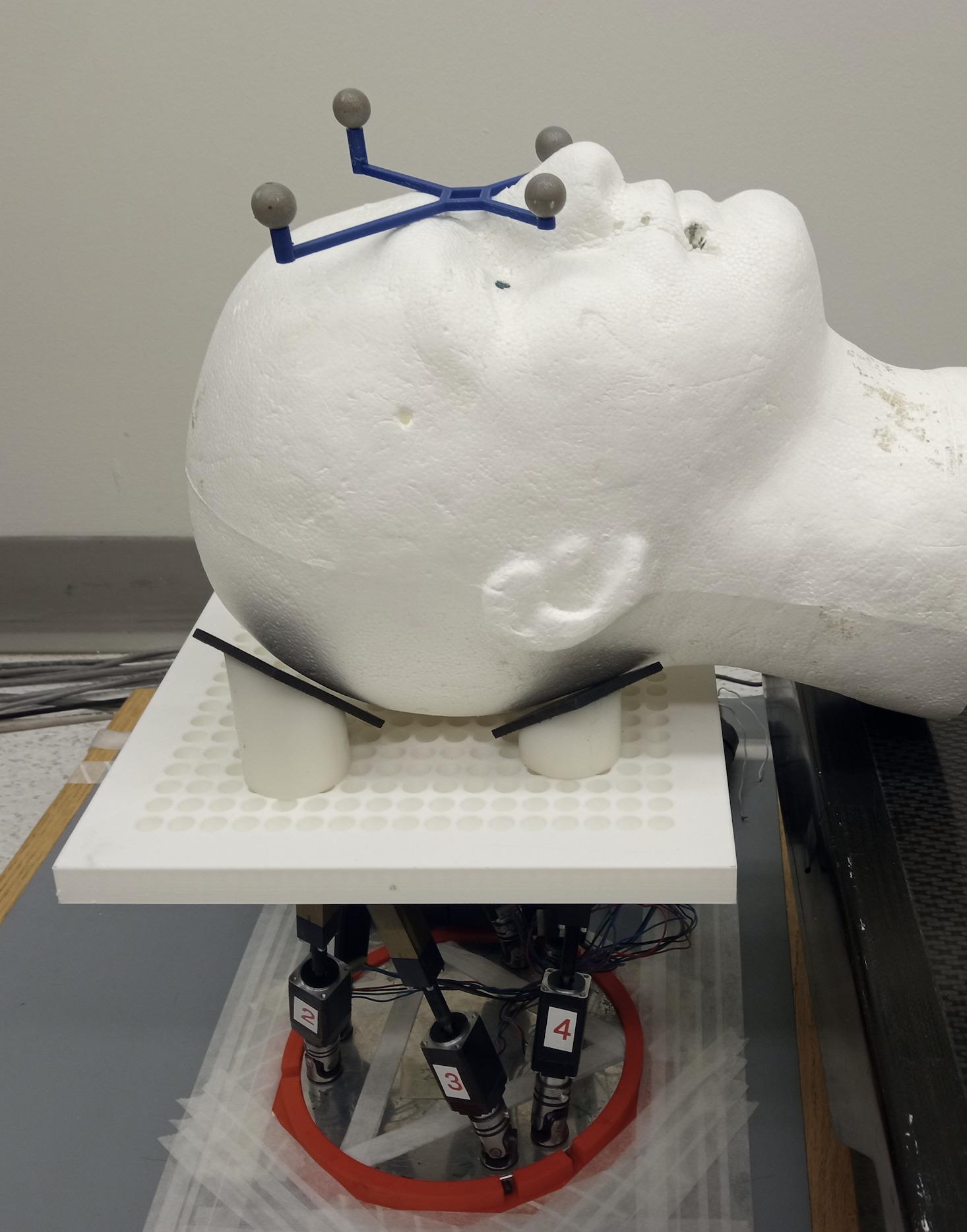}
	\includegraphics[height=60mm]{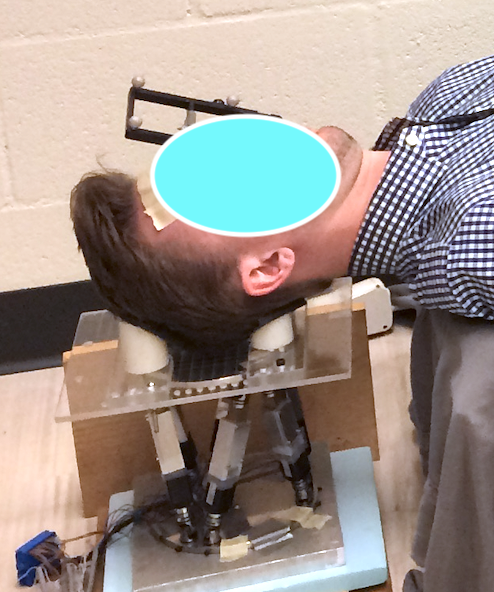}
	\caption{\footnotesize	
		A configuration of robotic head support with four plastic posts for facilitating head rotational control with \textcolor{black}{the} robot. A simple pillow support with only contact in the occipital region would not generate enough torque to rotate the head.
	}
	\label{headrest}
\end{figure*}

Real-time 6DoF head motion tracking is accomplished using an infrared (IR) stereoscopic camera (Polaris, NDI, Canada) with IR markers attached to the volunteer's head \cite{wiersma2013spatial}. Real-time 6DoF monitoring of head motion was performed at a 12 Hz sampling frequency. The head motion, \textcolor{black}{as tracked by} the four reflective IR markers, was subsequently transmitted automatically to an in-house developed LabVIEW (National Instruments) code\textcolor{black}{,} which controls the robotic device shown in the volunteer setup in Figure~\ref{hexapodSRS}. This control code communicates between the IR camera system and the robotic device simultaneously, transmitting the six linear actuator trajectories using an 8-axis stepper motor controller (PCI-7358, National Instruments) \cite{belcher2014development}. Movement of the robot produces a corresponding response motion in the volunteer's head by moving the cranial support structure (Fig.~\ref{headrest}). The tracked motion is then input into a control computer and analyzed using a fast iterative closest path control algorithm that is then sent to the robot to perform head positioning.

In this study, the head support consisted of four cushioned pads angled to help ensure contact with the volunteer\textcolor{black}{. These pads} were each connected to the robot's mobile top platform and \textcolor{black}{were} free to swivel around the AP axis to provide improved universal contact between volunteers while maintaining sufficient levels of comfort. The plastic posts were designed to be easy to install and remove, making \textcolor{black}{them} convenient for use with different head sizes \textcolor{black}{of volunteers}. The slope of the plastic posts was 45\textcolor{black}{$^\circ$}. Fig.~\ref{headrest} shows a schematic diagram of the head support system.

\subsection{Coordinate frames of LINAC, Target, Camera and IR tools}

\begin{figure*}[!t]
\centerline{
\includegraphics[height=35mm]{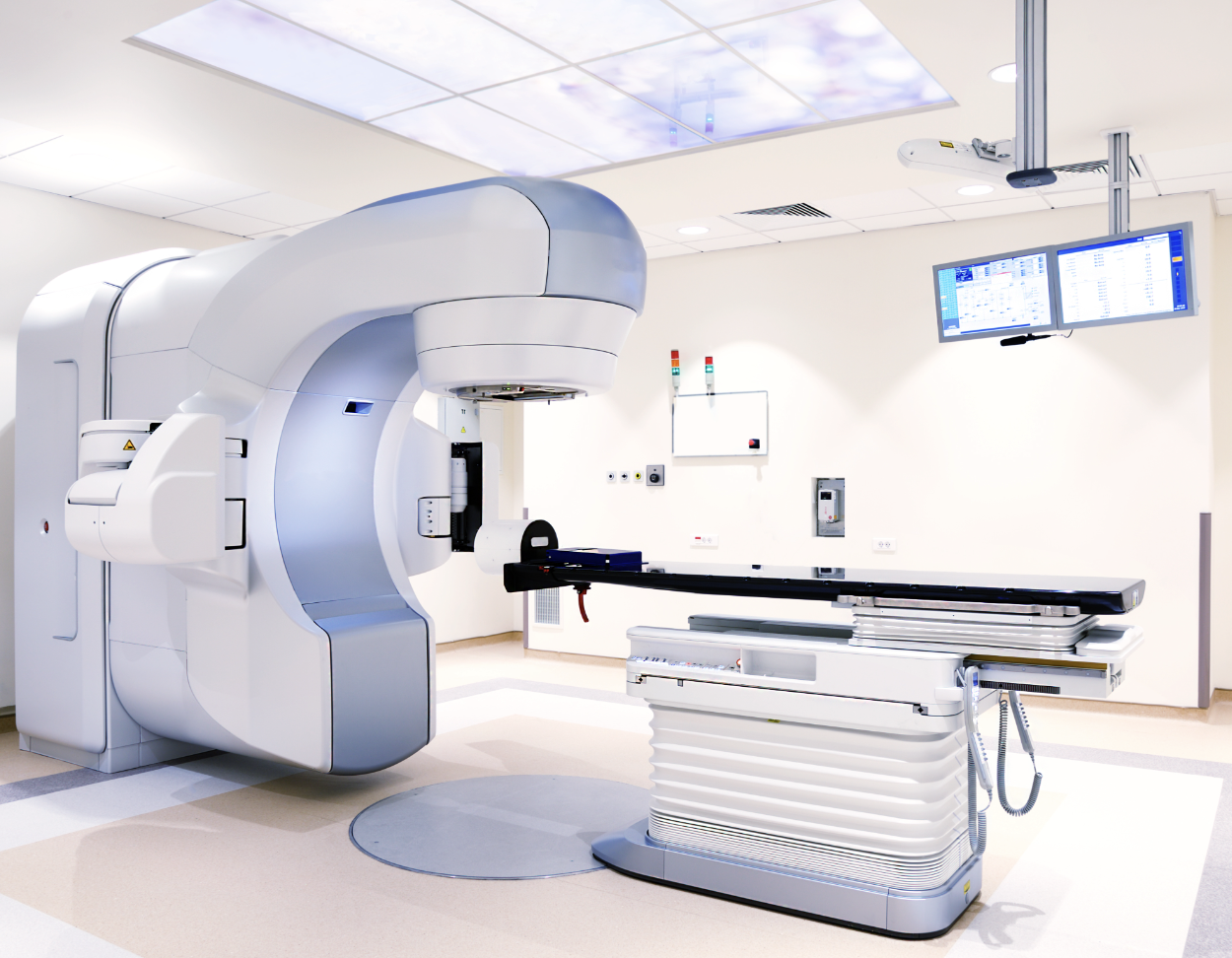}\hspace{3mm}
\includegraphics[height=35mm]{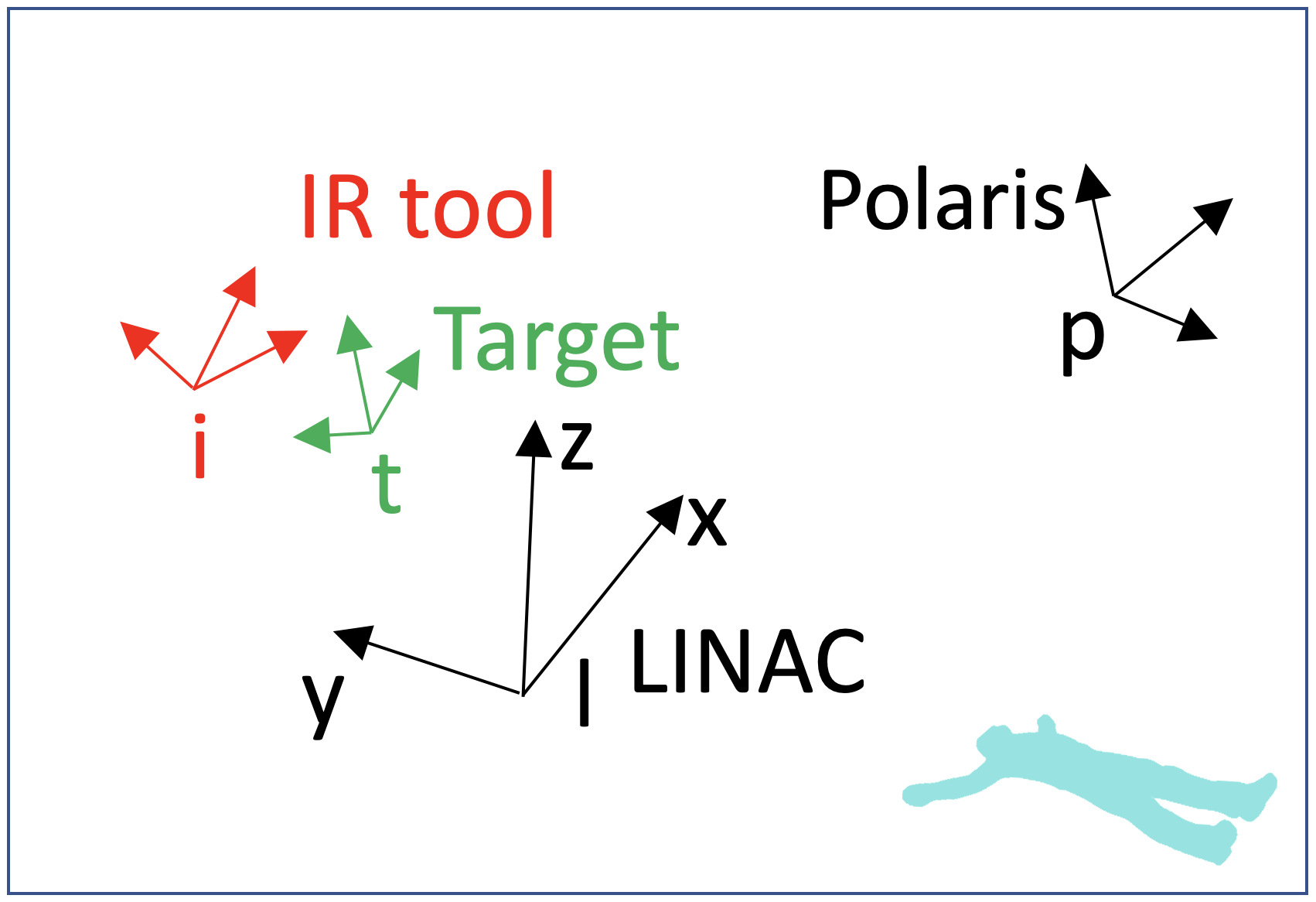}\hspace{3mm}
\includegraphics[height=35mm]{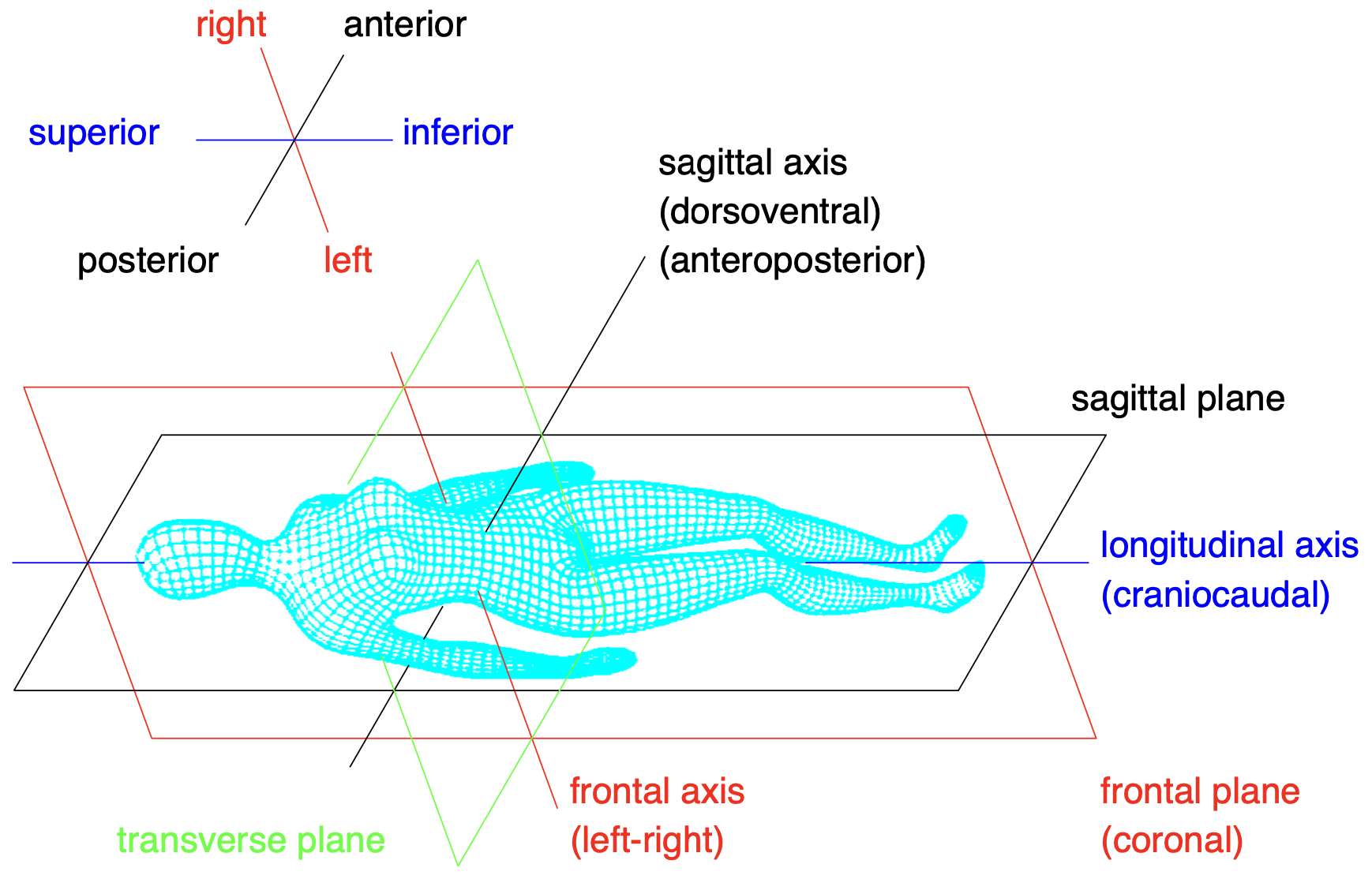}
}
\caption{
Different coordinate frames used for radiation delivering and position measuring for a patient in a LINAC system.
Left: A LINAC machine shown in a treatment room (image source: https://radformation.com).
Middle:  the coordinate frames related to the IR camera that tracks the patient's position are displayed. Right: the coordinate frame of a human lying on their back is defined.}
\label{LINAC_allCoords}
\end{figure*}

\textcolor{black}{
A medical linear accelerator (LINAC) is a device used to deliver external radiation beams for patient treatment. The LINAC coordinate system, with its origin at the isocenter, is defined to facilitate precise targeting. A mock tumor target’s 6D translational and rotational position within this system can be tracked using a camera system equipped with an IR marker tool (see Fig.~\ref{LINAC_allCoords}). Understanding the relationship between the four coordinate frames is crucial for enabling robotic systems to generate the required translational and rotational movements of a target point in the human head.
The following sub-sections outline key kinematic computations and control algorithms used in the robotic system. The following notation to specify the 6D position of the IR tool in the LINAC frame: ($R_l^i, \ell_l^i$). Here $R_l^i$ is the rotation matrix of the IR tool in LINAC frame (where the subscript $l$ represents LINAC frame, and the superscript $i$ represents the IR tool), and $\ell_l^i$ is the IR tool translational position (xyz) in LINAC frame.
}

\subsubsection{Calibrating the orientation of IR Camera to LINAC with translational motion}\label{cubeCali}
\begin{figure}[ht!]
\centerline{
\includegraphics[width=56mm]{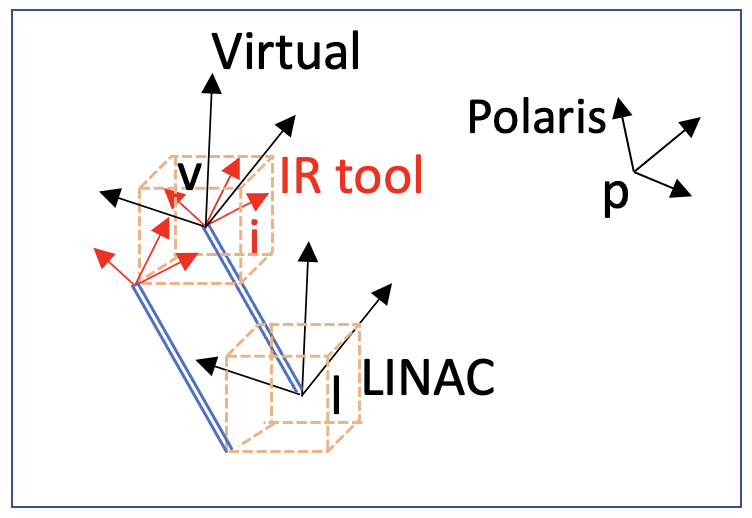}
}
\centerline{
\includegraphics[width=56mm]{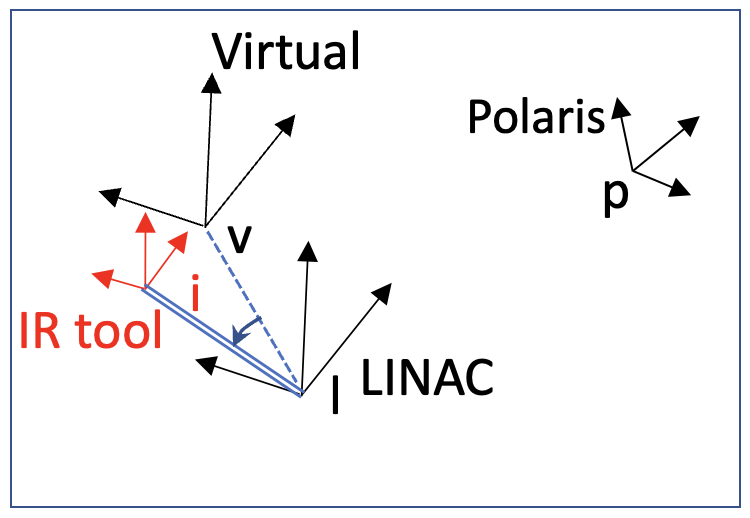}
}
\caption{Calibrating the IR Camera frame to the LINAC frame.
Top: calibrating the orientation of two frames with translational motion.
Bottom: determining translational offset of two frames with rotational motion.
}
\label{LINAC_coordsCubeMotion}
\end{figure}

One way of calibrating IR Camera and LINAC is rigidly attach an IR tool to the robotic platform, control the platform to make a pure translational motion in LINAC frame, and measure IR tool positions, see the top plot of Fig.~\ref{LINAC_coordsCubeMotion}.

Let the treatment table make a translational motion, and let the IR Camera record its positions at each point. Suppose these $m$ points in the LINAC frame are $L\in \bbR^{3\times m}$, and the corresponding IR Camera measurements are $P \in \bbR^{3\times m}$. 
For $L$ compute a mean vector $\bar l\in \bbR^{3\times 1}$, then horizontally concatenate $\bar l$ to generate a mean matrix $\bar L\in\bbR^{3\times m}$. Similarly, generate the mean vector $\bar p$ and the mean matrix $\bar P$ for the matrix $P$.
\[
C=L-\bar{L}, \quad D=P-\bar{P},
\]
\[
E=CD', \quad
[U, S, V]= svd (E)
\]
\[
R_l^p=U \left[ \begin{matrix} 1&0&0\cr 0&1&0 \cr 0&0&\mbox{det} (UV') \end{matrix} \right] V'
\]

\textcolor{black}{
A virtual coordinate frame can be defined in which the orientation of the xyz axes matches that of the LINAC frame, but the origin is positioned at the virtual point (origin) of the IR tool when the treatment table is at the isocenter.}
That is, $R_v^p = R_l^p$.
The vector $\bfv_v^{vp}$ in the virtual frame is $\ell_v^p=\bar l- R_l^p \bar p$.

For IR tool position specified by ($R_p^i, \ell_p^i$) in IR Camera frame,
it can be specified by ($R_v^i, \ell_v^i$) in the virtual frame as following, 
\[
R_v^i=R_v^p R_p^i=R_l^p R_p^i,
\]
\[
\ell_v^i 
= R_v^p (\ell_p^i - \ell_p^v) 
= R_l^p (\ell_p^i - \ell_p^v) 
= R_l^p  \ell_p^i + \ell_v^p
\]

Align the target to LINAC frame, and consider this is target zero position in LINAC frame. Record the initial 
($\bar R_v^i, \bar\ell_v^i$).
The target position in Virtual frame
($R_v^t , \; \ell_v^t$) can be calculated by
$
R_v^t = R_v^i (\bar R_v^i)',\quad
\ell_v^t = \ell_v^i -R_v^t\bar\ell_v^i
$.

\subsubsection{Computing the coordinates of the virtual point of the IR tool within the LINAC frame}\label{calcvirtual}

\textcolor{black}{
Determining the coordinates of the virtual point (denote as $\ell_l^v$) of the IR tool in the LINAC frame is an important intermediate step for computing the target position relative to the robot, see the bottom plot of Fig.~\ref{LINAC_coordsCubeMotion}. To obtain these coordinates, we perform pure rotational motions around the origin of the LINAC frame, and record corresponding
($R_v^t, \ell_v^t$).
}
For each rotation
$
R_v^t \ell_l^v -\ell_l^v= \ell_v^t.
$
Suppose $n$ pure rotations are performed. Vertically concatenate all $(R_v^t - I_3)$ and obtain a matrix $A\in\bbR^{3n\times 3}$, and similarly vertically concatenate all $\ell_v^t$ and obtain a $b\in\bbR^{3n\times 1}$. Then
$
\ell_l^v = A\backslash b.
$
For IR tool position specified by ($R_p^i, \ell_p^i$) in IR Camera frame,
it can be specified by ($R_l^i, \ell_l^i$) in LINAC frame as following, 
\begin{equation}
\label{equ:LF}
R_l^i=R_v^i, \quad
\ell_l^i = \ell_v^i+\ell_l^v
\end{equation}

\subsubsection{Calibrate IR Camera to LINAC}

With the algorithms in the subsections ~\ref{cubeCali} and \ref{calcvirtual}, IR camera coordinate frame can be calibrated to LINAC frame.
For IR tool position specified by ($R_p^i, \ell_p^i$) in IR Camera frame,
it can be specified by ($R_l^i, \ell_l^i$) in LINAC as following, 
\begin{equation}
\label{LINAC}
R_l^i=R_l^p R_p^i, \quad
\ell_l^i 
= R_l^p  \ell_p^i + \ell_v^p + \ell_l^v
\end{equation}
\textcolor{black}{
Since the orientation of the Virtual frame and LINAC is the same, one has $\ell_l^p=\ell_v^p+\ell_l^v$.
}

\subsubsection{
\textcolor{black}{
Tracking target}
}

\begin{figure}[ht]
\centerline{
\includegraphics[width=56mm]{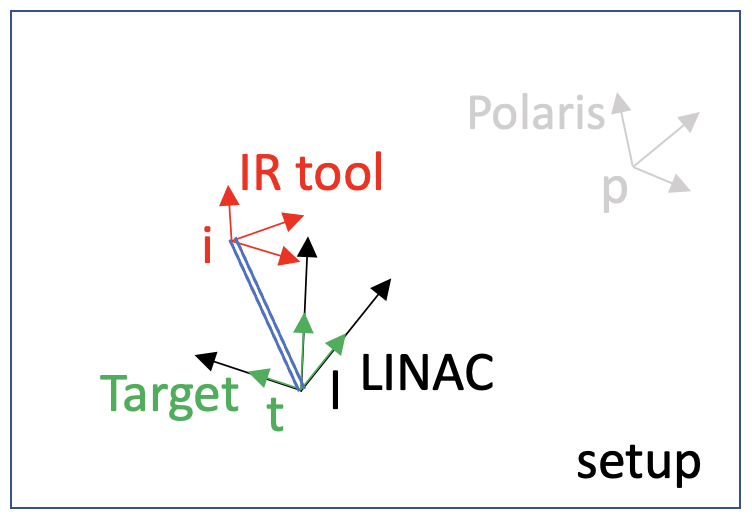}
}
\centerline{
\includegraphics[width=56mm]{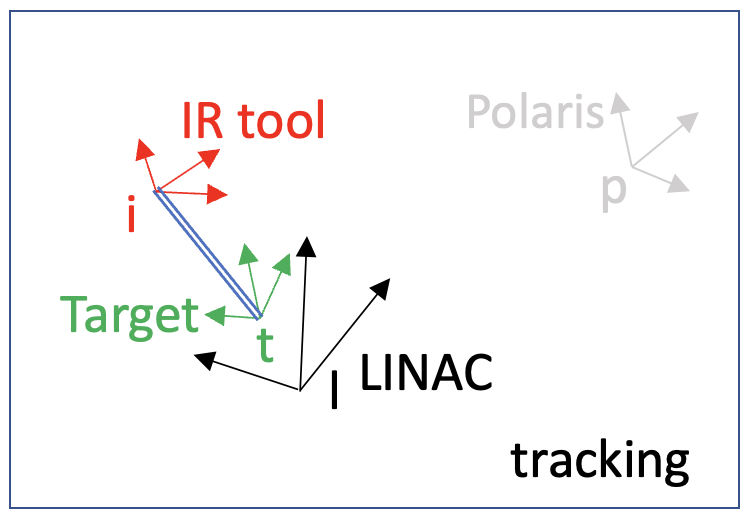}
}
\caption{
\textcolor{black}{
Target tracking by using IR tool.
Top: During volunteer setup, the mock target on the volunteer's head is aligned with the LINAC frame.
Bottom: The displacement of the target is monitored in the LINAC frame using the IR tool’s 6D position tracking.}
}
\label{LINAC_coordsTracking}
\end{figure}

\textcolor{black}{
IR tool was rigidly attached to a volunteer head for position tracking.
During setup, as shown in the top plot of Fig.~\ref{LINAC_coordsTracking}, one can first align the orientation and origin of a mock target within a volunteer’s head to the LINAC frame, then record the IR tool position at that moment, and denote as ($\bar R_l^i, \bar \ell_l^i$).
During target tracking as in the bottom plot of Fig.~\ref{LINAC_coordsTracking},
\[
R_l^i = R_l^t R_t^i, \quad \ell_l^i 
= \bfv_l^{lt}+ \bfv_l^{ti}
= \ell_l^t+ R_l^t \ell_t^i
\]
where $(R_t^i, \ell_t^i)$ represent the IR 6D position in the target frame. Since target and IR tool are rigid connected,
($R_t^i, \ell_t^i$) = ($\bar R_l^i, \bar \ell_l^i$).
Then the target position in LINAC frame, ($R_l^t, \ell_l^t$), can be calculated by
$
R_l^t=R_l^i (\bar R_l^i)', \quad
\ell_l^t= \ell_l^i - R_l^t \bar \ell_l^i
$.
}

\subsection{Volunteer Experiments}

\textcolor{black}{
Human and robot interactions were studied using three volunteers (IRB \#843058). During the study, volunteers assumed a relaxed supine position on the simulated treatment table, with their heads supported by the robot.  Volunteers were requested to lie down with their heads positioned to maximize contact with four plastic posts in the robotic head support. The robot then performed serial translational and rotational motions, while the IR camera recorded the resultant head movements. To study human -robot reaction thoroughly, the robot was programmed to move along the frontal axis (left-right, defined as the x axis), the longitudinal axis (superior-inferior, defined as the y axis) and the sagittal axis (anterior-posterior, defined as the z axis) $\pm 7$mm, $\pm 4$mm and $\pm 7$mm respectively. The robot was also programmed to pitch (around the frontal axis), roll (around the longitudinal axis) and yaw (around the sagittal axis) $\pm 5$ degrees, $\pm 5$ degrees and  $\pm 3$ degrees respectively. Rotation occurred around a mock target within the volunteer's head. Two mock target positions were considered, one at cervical vertebrae C1, and the other at a point displaced (-60, 120, 60) mm from the C1, i.e., at a position 60 mm to the right, 120mm superiorly and 60mm anteriorly with respect to C1.
The 6D head motion of each volunteer was recorded and analyzed. Motion limits were evaluated based on head-neck biomechanics to determine physiological constraints.
}

\begin{figure}[ht!]
\centerline{
\includegraphics[width=60mm]{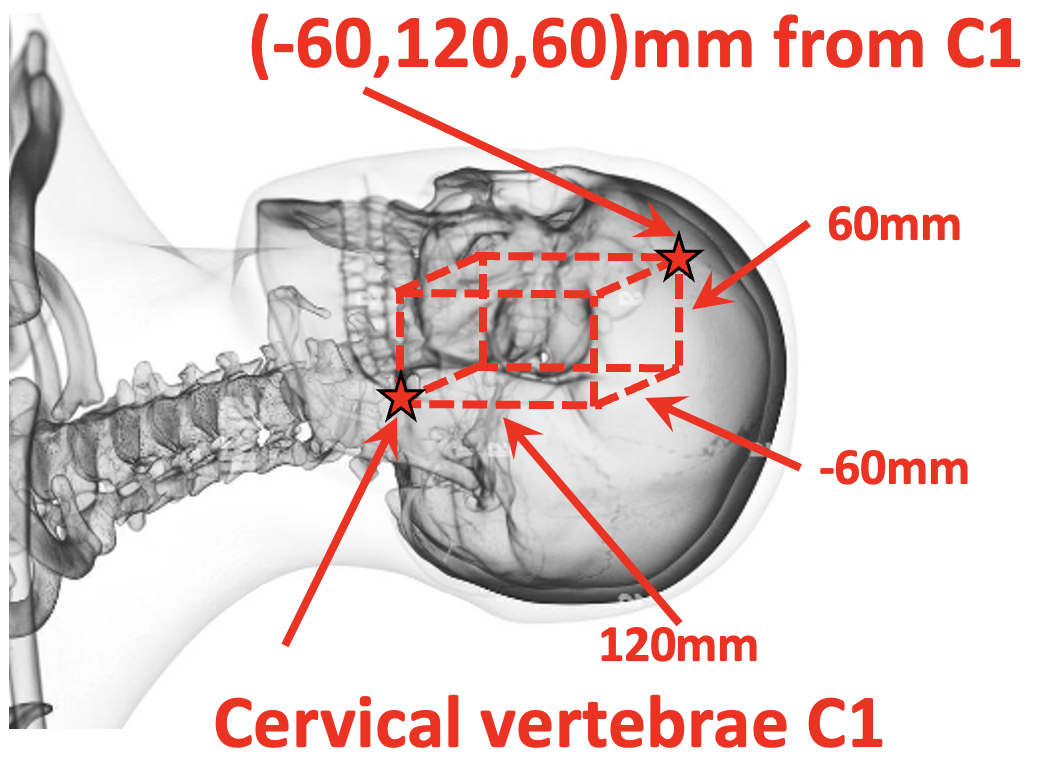}
}
\caption{Robot rotated around different target points in human head}
\label{fig_rotatingPoints}
\end{figure}

\section{Results}
\label{results}

\textcolor{black}{
As described in the methods section, a coordinate calibration procedure was conducted to align the coordinate frames of the IR camera and the LINAC system. To assess the accuracy of this calibration, a head phantom with attached IR markers was used alongside virtual targets positioned within the brain (Fig.~\ref{headrest}). The virtual targets included cervical vertebrae C1 and a second target located at $(-60, 120, 60)$ mm relative to C1, as shown in Fig.~\ref{fig_rotatingPoints}.
The targets were programmed to execute controlled motion along three axes: frontal ($\pm 7$ mm), longitudinal ($\pm 4$ mm), and sagittal ($\pm 7$ mm). Additionally, rotational movements were introduced around these axes, with rotations of $\pm 5^\circ$, $\pm 5^\circ$, and $\pm 3^\circ$, respectively, achieved through robotic actuation. The ranges of motion were selected based on the requirements of SRS applications, reflecting typical displacement magnitudes encountered during treatment to ensure clinical relevance.
Head and neck anatomical constraints were also taken into account. For example, the superior-inferior (y-axis) direction exhibits limited displacement due to anatomical rigidity. As a result, a smaller range was chosen for this axis compared to the frontal (x-axis) and sagittal (z-axis) directions, which allow for greater mobility.
}

\begin{figure*}[!t] 
	\centering
	\includegraphics[width=106mm]{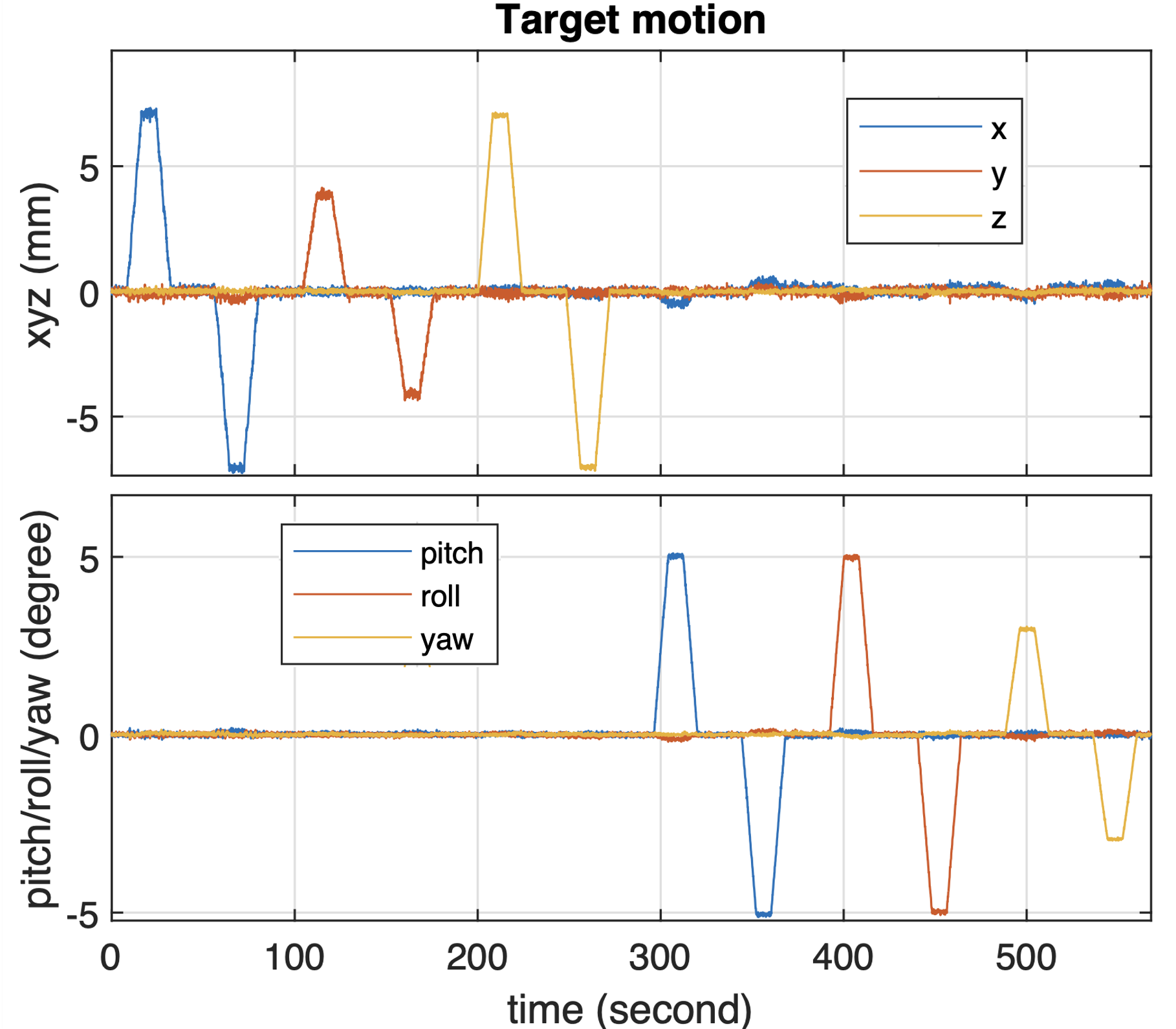}
	\caption{\footnotesize
The measured virtual target cervical vertebrae C1 trajectory followed the desired pure translational and rotational motion.
	}
	\label{fig_WS_xyzpsi}
\end{figure*}

The resulting trajectory of cervical vertebrae C1, captured in Fig.~\ref{fig_WS_xyzpsi}, demonstrated a close adherence to the intended motion, due to the rigid coupling between the phantom and the robotic system. 
A comparison between the measured and expected target trajectories revealed a strong agreement, yielding a root mean square error (RMSE) of 0.3 mm in positional accuracy and 0.2$^\circ$ in angular precision. Similar results were obtained for the virtual target located at $(-60, 120, 60)$ mm from C1, further validating the robustness of the calibration approach.

For humans, it is not expected that the head will exactly follow the robot motion due to the anatomical constraints imposed by the cervical vertebrae (C7) and thoracic vertebra (T1), as well as the connection between the volunteer’s body and the treatment table. The biomechanical properties of the head-neck system introduce additional complexity, where the intervertebral joints, ligamentous structures, and muscle tone contribute to motion coupling across multiple axes \cite{swartz2005cervical,yoganandan2001biomechanics}.

\textcolor{black}{
To investigate this, human volunteers were asked to lay down in a relaxed supine position on the treatment table with their head comfortably supported by the robotic head holder (Fig.~\ref{headrest}). Head motion was then recorded while the robot performed isolated x, y, z translations, as well as pitch, roll, and yaw rotations. 
Fig.~\ref{fig_WS_oneone_tran_rota} plots the head response of three sample volunteers to these motions. As observed, head motion does not precisely follow the robot’s input due to constraints imposed by the head-neck system. Several key biomechanical features emerged across all volunteers:
\begin{enumerate}
    \item Head movement patterns were consistent across trials, indicating a systematic response rather than random variation.
    \item The superior-inferior (y) direction exhibited the least motion range, likely due to the stability provided by surrounding musculature and the articulation of the lower cervical spine.
    \item Motion coupling between different dimensions was consistently observed, highlighting the influence of neck joint biomechanics and passive constraints on translational and rotational head movements.
\end{enumerate}
}

These findings indicate that head motion under external robotic forces is governed by anatomical constraints and biomechanical coupling effects, which should be taken into account in precision-based applications involving head stabilization or movement prediction.

To examine the motion ranges of the human head workspace, Fig.~\ref{WS_box_oneone} plots the robot's range versus \textcolor{black}{the} induced motion of the head for all three volunteers. For \textcolor{black}{translational} motion, in all cases \textcolor{black}{the study} found that the left-right (x) and \textcolor{black}{anterior-posterior} (z) \textcolor{black}{directions} had significant head motion ranges \textcolor{black}{of} $>10$ mm, \textcolor{black}{while} the superior-inferior (y) \textcolor{black}{direction} had \textcolor{black}{a} limited range of $<5$ mm. For rotational motion, head roll had the largest motion range, followed by pitch and yaw. A summary of the results \textcolor{black}{is} given in Table~\ref{tab:myfirsttable}.

\textcolor{black}{
In all cases, human head motion was found to be coupled with other dimensions during robot-assisted guidance. As shown in Fig.~\ref{WS_coupling} for a sample volunteer, coupling can be seen in one or more different dimensions in response to robot motion along a particular direction. For instance, when the robot moves along the left-right axis, the volunteer's reaction occurs primarily along the left-right direction, with coupled (about 1.5 degrees) rotation around the sagittal axis, see the first row in Figure 11-A. 
Similarly, motion along the sagittal axis (anterior-posterior) predominantly elicited sagittal displacement, with additional coupled pitch (about 2 degrees) around the frontal axis,
see the third row in Figure 11-A.
}

In addition to the mock cervical vertebrae C1 target (\textcolor{black}{denoted} as target `A'), \textcolor{black}{the} volunteer head motion range was also explored at another point within the brain\textcolor{black}{,} located \textcolor{black}{at a position} 60~mm \textcolor{black}{to the} right, 120~mm \textcolor{black}{superior to}, and 60~mm \textcolor{black}{anterior to} \textcolor{black}{the} C1 \textcolor{black}{landmark} (\textcolor{black}{denoted} as target `B'). This \textcolor{black}{analysis is} important since SRS radiation targets can be located \textcolor{black}{in} different \textcolor{black}{regions of} the brain\textcolor{black}{,} and it is important to know \textcolor{black}{the} expected motion range of \textcolor{black}{a} target. Similar to target A, the robot performed \textcolor{black}{the} similar motion trajectories around target B\textcolor{black}{,} and the results \textcolor{black}{were} analyzed. Fig.~\ref{WS_box_positions} compares the \textcolor{black}{volunteers'} reaction \textcolor{black}{ranges} for motion around targets A and B\textcolor{black}{,} respectively. \textcolor{black}{The results show that} the motion range \textcolor{black}{varies} depending on \textcolor{black}{the target's location within} the brain.

\section{Discussion}
\label{discussion}

This work represents the first \textcolor{black}{exploration of} the 6DoF interaction workspace for \textcolor{black}{robot-assisted} head motion control \textcolor{black}{in} human test subjects. It provides a deeper understanding of human anatomy by utilizing a unique modeling technique that can be used for various applications. Unlike the majority of head and neck modeling studies \textcolor{black}{that} focus on high-impact \textcolor{black}{and large-motion} simulations for accident research \textcolor{black}{found} in the literature, our work introduced a novel approach \textcolor{black}{using} high-precision motion measurements \textcolor{black}{to explore} the natural motion ranges of the human head. These results \textcolor{black}{thus} provide useful data \textcolor{black}{for} defining fundamental motion constraints and trajectory information \textcolor{black}{applicable to developing} accurate biomechanical \textcolor{black}{models of} the head\textcolor{black}{-}neck relationship.

\begin{figure*}[!t] 
	\centering
	\includegraphics[width=130mm]{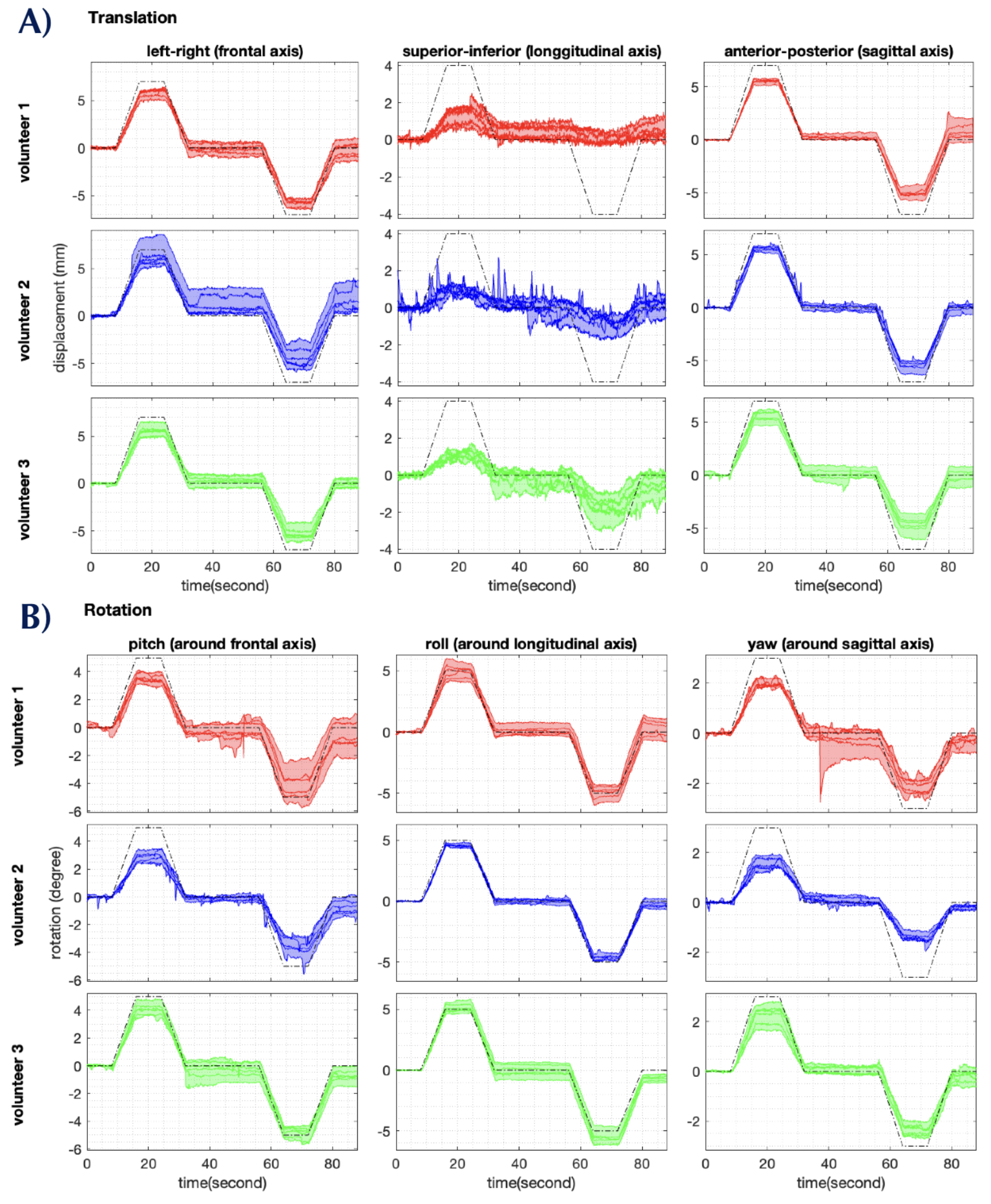}
	\caption{\footnotesize
		Volunteer head reaction to robot translational and rotational motion. A) corresponded to one volunteer with different motion type (left-right, superior-inferior, and anterior-posterior) indicated by columns. Along each direction, the particular robot motion was repeated three times. 
Similarly, B) corresponded to one volunteer with different motion type (pitch, roll and yaw).
	}
	\label{fig_WS_oneone_tran_rota}
\end{figure*}

\begin{figure*}[!t] 
	\centering
	\includegraphics[width=160mm]{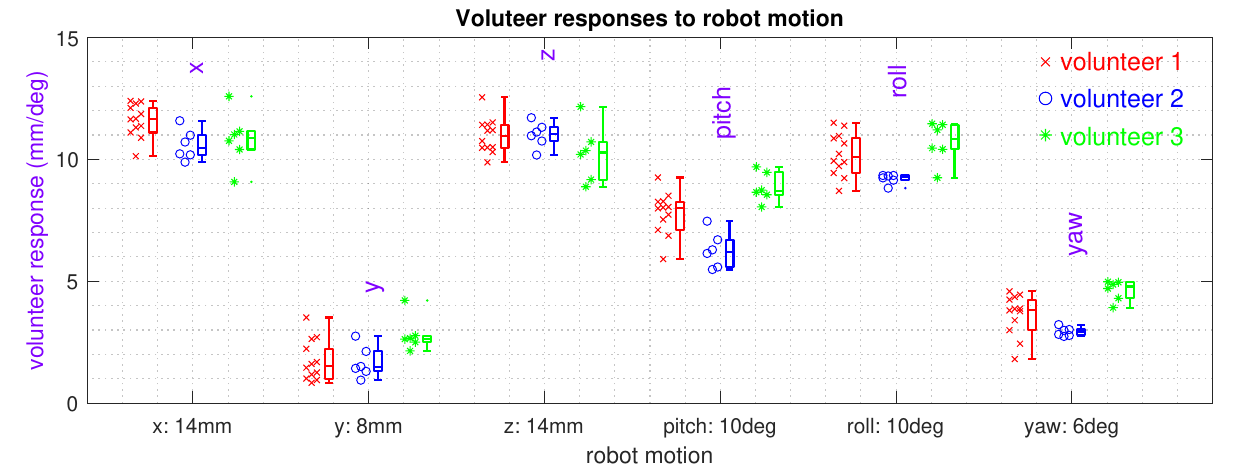}
	\caption{\footnotesize
		Boxplot of the ranges of correponding volunteers reaction to individual robot motion.
	}
	\label{WS_box_oneone}
\end{figure*}

\begin{figure*}[!t] 
\centering
\includegraphics[width=160mm]{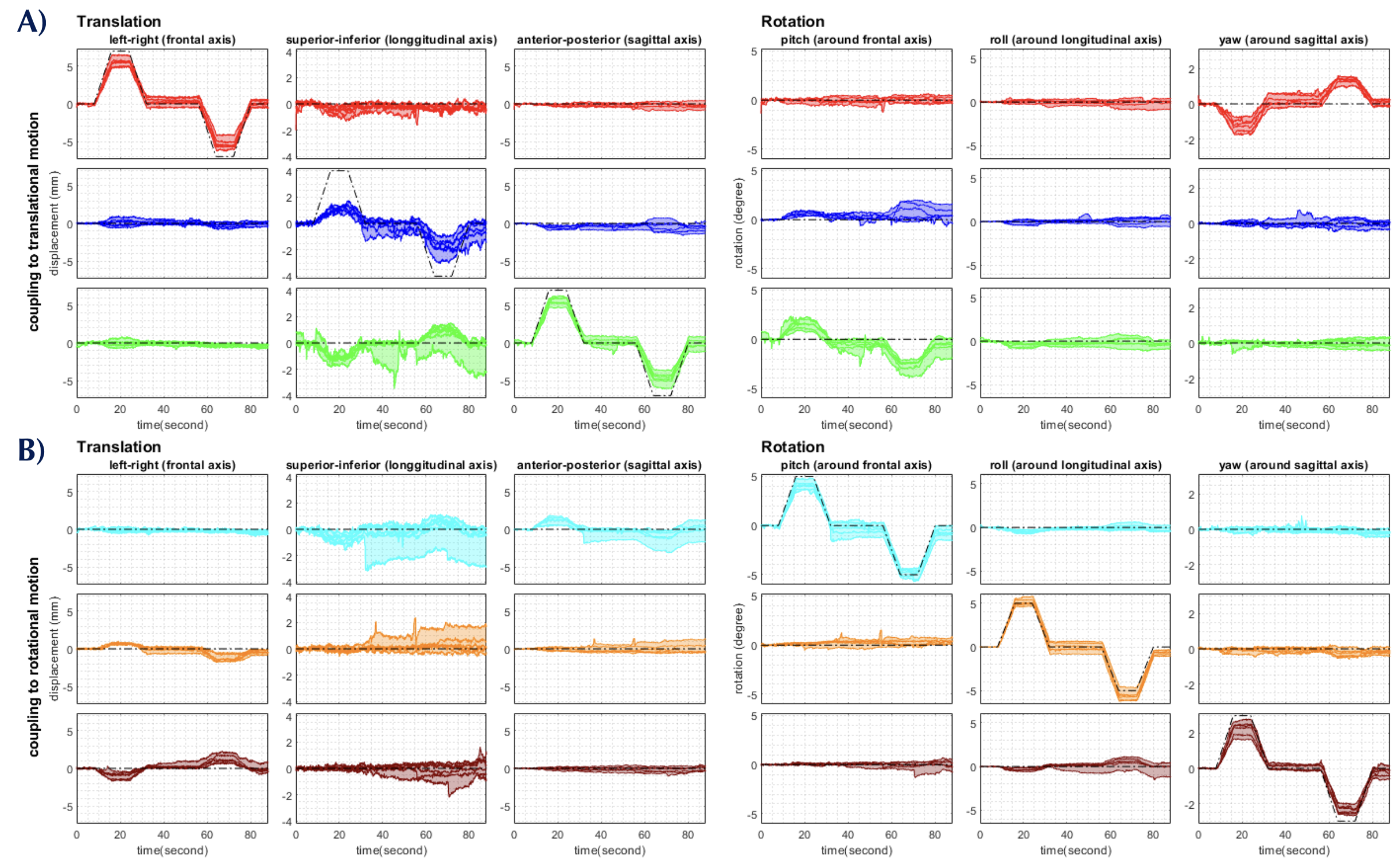}
\caption{\footnotesize
Sample data for one volunteer displaying \textcolor{black}{dimensional coupling in} head motion \textcolor{black}{resulting} from robot motion along a particular direction. \textcolor{black}{(A)} \textcolor{black}{shows} the volunteer's reaction to translational robot motion, \textcolor{black}{while (B)} \textcolor{black}{shows} reaction to rotational motion.
}
\label{WS_coupling}
\end{figure*}

\begin{figure*}
	\centering
	\includegraphics[width=160mm]{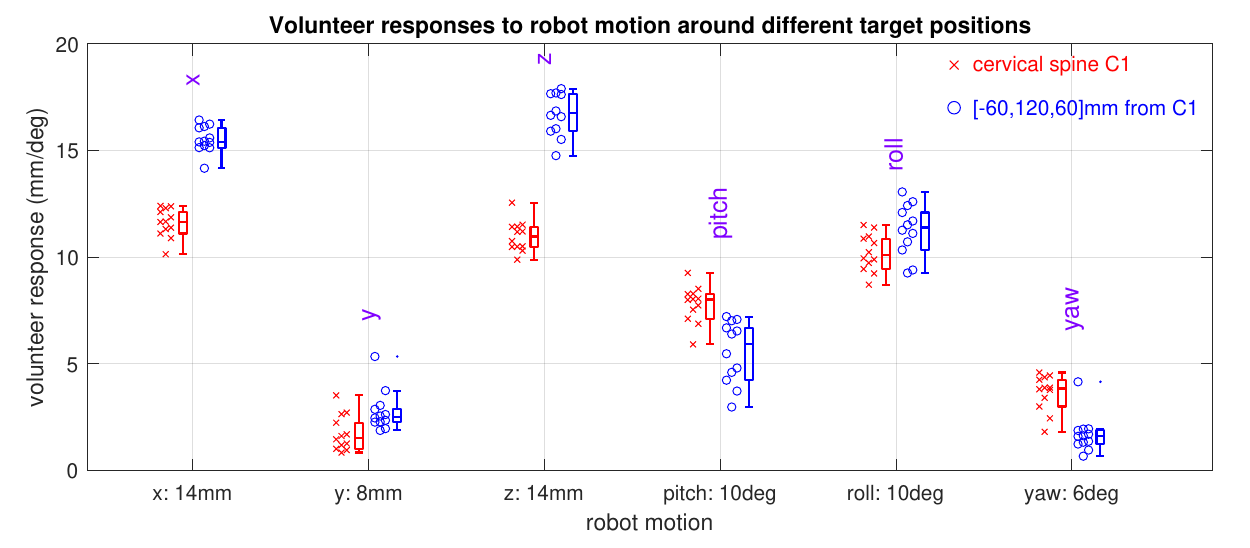}
	\caption{\footnotesize
		Human motion around different tumor target positions. One was in cervical vertebrae C1, and the other was [-60mm, 120mm, 60mm] away from C1. The left boxplot in each pair was corresponding to robot motion moving around C1.
	}
	\label{WS_box_positions}
\end{figure*}

Although the volunteers \textcolor{black}{had} different heights, weights, \textcolor{black}{genders}, and other characteristics, \textcolor{black}{we} found that \textcolor{black}{the} robotically explored head motion workspace was surprisingly similar. As shown in Fig.~\ref{WS_box_oneone}, for each robot-induced motion along a particular direction\textcolor{black}{,} the head motion response was similar across all volunteers. Additionally, \textcolor{black}{we} found that the head motion response was reproducible for the same robot motion trajectory\textcolor{black}{,} as shown by the individual points in Fig.~\ref{WS_box_oneone}. This similarity and reproducibility \textcolor{black}{across} different volunteers \textcolor{black}{is} very encouraging for biokinematic modeling \textcolor{black}{as it suggests that a} semi-accurate universal model could be constructed \textcolor{black}{using} nine rigid bodies\textcolor{black}{,} representing the head, cervical vertebrae C1-C7, and thoracic vertebra T1. \textcolor{black}{The} real-time head motion trajectory data shown in Fig.~\ref{fig_WS_oneone_tran_rota} and Fig.~\ref{WS_coupling} could be used for model testing.

\textcolor{black}{A} central limitation \textcolor{black}{for} motion control was not the robot\textcolor{black}{,} but rather the motion limits of the head\textcolor{black}{-}neck \textcolor{black}{system}. The kinematic data in Table~\ref{tab:myfirsttable} show\textcolor{black}{} that\textcolor{black}{,} regardless of volunteer \textcolor{black}{characteristics}, the most limited motion range \textcolor{black}{occurred} along the superior-inferior direction\textcolor{black}{,} with \textcolor{black}{a magnitude of} $2.0\pm 0.6$ mm. The second most limited motion range was \textcolor{black}{for} yaw rotation\textcolor{black}{,} at $3.7\pm 0.5$\textcolor{black}{$^\circ$}. For high-accuracy head motion control in robotic SRS treatment\textcolor{black}{,} the limited range along the superior-inferior direction could pose \textcolor{black}{a challenge} if the starting position of the surgical target \textcolor{black}{lies} outside this range. \textcolor{black}{We} propose a simple pre-treatment test where the robot \textcolor{black}{first executes} a sequence of well-defined moves along various 6D axes to probe the patient's workspace. Once the interaction workspace is determined\textcolor{black}{,} the robot can \textcolor{black}{then position} the head in the center of this workspace to allow for the greatest range of motion during \textcolor{black}{the} actual robotic SRS treatment.

\textcolor{black}{
the observed small displacement in the superior-inferior direction due to the bio-kinematics of vertebrae, provide valuable insights. This finding has proven to be crucial in our ongoing patient trial treatments with the robotic motion compensation system. It has directly influenced the design of the head holder within our system, ensuring that it accommodates and leverages the anatomical limitations in the superior-inferior direction. Additionally, this understanding has informed how we set up patients, taking the constrained displacement into account to optimize the system's performance.
}

\textcolor{black}{
This study demonstrates that the robotic biokinematic assessment platform is a promising tool for evaluating head-neck stabilization. While the small sample size limits broad conclusions, the consistency in repeated trials shows that the method produces reliable data. These early results suggest that even with a small group, meaningful insights can be gained. However, to strengthen the findings and improve clinical relevance, we have launched a larger study with 40 participants (20 healthy volunteers and 20 clinical patients). This expanded research will perform more detailed statistical analyses and further refine the platform for future applications.
}

\section{Conclusions}
\label{conclusions}

\textcolor{black}{
This study introduced a novel method for studying human head workspace in the supine position. Using a robot loosely coupled to the back of the head, we explored the mechanics of the head and neck, laying the groundwork for future biokinematic modeling studies. Our findings show that robot-induced motion in a specific direction led to consistent head motion responses across all volunteers. These responses were also reproducible when the same robotic motion trajectories were applied, confirming the reliability of this approach.
As a preliminary study, the focus was on assessing the feasibility and consistency of this robotic biokinematic platform. This research is an important first step in studying head-neck biokinematics within the context of robotic stabilization. Future work will refine the methods and expand the dataset to better address the complexities of head-neck mechanics.
}

\section*{Acknowledges}

Research reported in this publication was supported by the National Institutes of Health (NIH) under award number R01CA227124.

\begin{table*}
  \caption{Volunteers motion range with respect to robot motion (x:14mm, y:8mm, z:14mm, pitch:10deg, roll:10deg and yaw:6deg)}
  \centering
  \begin{tabular}{|c|c|c|c|c|c|c|}
\hline
& x&	y&	z&	pitch&	roll&	yaw\\  
\hline
Volunteer1&	11.60$\pm$0.68&	1.75$\pm$0.84&	10.97$\pm$0.72&	7.79$\pm$0.86&	10.21$\pm$0.88&	3.63$\pm$0.85 \\ 
\hline
Volunteer2&	10.60$\pm$0.63&	1.67$\pm$0.65&	11.01$\pm$0.52&	6.27$\pm$0.74&	9.20$\pm$0.20&	2.93$\pm$0.18\\
\hline
Volunteer3&	10.86$\pm$0.40&	2.47$\pm$0.31&	10.56$\pm$1.51&	8.75$\pm$0.71&	10.30$\pm$0.99&	4.40$\pm$0.54\\
\hline
  \end{tabular}
  \label{tab:myfirsttable}
\end{table*}

\bibliographystyle{IEEEtran}
\bibliography{references}

@article{gaspar1997recursive,
  title={Recursive partitioning analysis (RPA) of prognostic factors in three Radiation Therapy Oncology Group (RTOG) brain metastases trials.},
  author={Gaspar, Laurie and Scott, Charles and Rotman, Marvin and Asbell, Sucha and Phillips, Theodore and Wasserman, Todd and McKenna, W Gillies and Byhardt, Roger},
  journal={International journal of radiation oncology, biology, physics},
  volume={37},
  number={4},
  pages={745--751},
  year={1997}
}

@article{belcher2017towards,
  title={Towards frameless maskless SRS through real-time 6DoF robotic motion compensation},
  author={Belcher, Andrew H and Liu, Xinmin and Chmura, Steven and Yenice, Kamil and Wiersma, Rodney D},
  journal={Physics in Medicine \& Biology},
  volume={62},
  number={23},
  pages={9054},
  year={2017},
  publisher={IOP Publishing}
}

@article{wiersma2019conceptual,
  title={A conceptual study on real-time adaptive radiation therapy optimization through ultra-fast beamlet control},
  author={Wiersma, Rodney D and Liu, Xinmin},
  journal={Biomedical physics \& engineering express},
  volume={5},
  number={5},
  pages={055016},
  year={2019},
  publisher={IOP Publishing}
}

@article{salem2013vivo,
  title={In vivo three-dimensional kinematics of the cervical spine during maximal axial rotation},
  author={Salem, Walid and Lenders, Cyrille and Mathieu, Jacques and Hermanus, Nicole and Klein, Paul},
  journal={Manual therapy},
  volume={18},
  number={4},
  pages={339--344},
  year={2013},
  publisher={Elsevier}
}

@article{jalali2009micromultileaf,
  title={Micromultileaf collimator-based stereotactic radiosurgery for selected arteriovenous malformations: Technique and preliminary experience},
  author={Jalali, Rakesh and Dutta, Debnarayan and Srinivas, C and Munshi, Anusheel and Limaye, Uday and Goel, A and Deshpande, Deepak and Sarin, Rajiv and others},
  journal={Journal of Cancer Research and Therapeutics},
  volume={5},
  number={3},
  pages={186},
  year={2009},
  publisher={Medknow Publications}
}

@article{betti1989stereotactic,
  title={Stereotactic radiosurgery with the linear accelerator: treatment of arteriovenous malformations},
  author={Betti, Osvaldo O and Munari, Claudio and Rosler, Roberto},
  journal={Neurosurgery},
  volume={24},
  number={3},
  pages={311--321},
  year={1989},
  publisher={Oxford University Press}
}

@article{chen2004treatment,
  title={Treatment of trigeminal neuralgia with linear accelerator radiosurgery: initial results},
  author={Chen, Joseph CT and Girvigian, Michael and Greathouse, Hugh and Miller, Michael and Rahimian, Javad},
  journal={Journal of neurosurgery},
  volume={101},
  number={Supplement3},
  pages={346--350},
  year={2004},
  publisher={American Association of Neurological Surgeons}
}

@article{zahra2009stereotactic,
  title={Stereotactic radiosurgery for trigeminal neuralgia utilizing the BrainLAB Novalis system},
  author={Zahra, Hadi and Teh, Bin S and Paulino, Arnold C and Yoshor, Daniel and Trask, Todd and Baskin, David and Butler, E Brian},
  journal={Technology in cancer research \& treatment},
  volume={8},
  number={6},
  pages={407--412},
  year={2009},
  publisher={SAGE Publications Sage CA: Los Angeles, CA}
}

@article{litre2013fractionated,
  title={Fractionated stereotactic radiotherapy for acoustic neuromas: a prospective monocenter study of about 158 cases},
  author={Litre, Fabien and Rousseaux, Pascal and Jovenin, Nicolas and Bazin, Arnaud and Peruzzi, Philippe and Wdowczyk, Didier and Colin, Philippe},
  journal={Radiotherapy and Oncology},
  volume={106},
  number={2},
  pages={169--174},
  year={2013},
  publisher={Elsevier}
}

@article{murphy2003patterns,
  title={Patterns of patient movement during frameless image-guided radiosurgery},
  author={Murphy, Martin J and Chang, Steven D and Gibbs, Iris C and Le, Quynh-Thu and Hai, Jenny and Kim, Daniel and Martin, David P and Adler Jr, John R},
  journal={International Journal of Radiation Oncology* Biology* Physics},
  volume={55},
  number={5},
  pages={1400--1408},
  year={2003},
  publisher={Elsevier}
}

@article{murphy2009intrafraction,
  title={Intrafraction geometric uncertainties in frameless image-guided radiosurgery},
  author={Murphy, Martin J},
  journal={International Journal of Radiation Oncology* Biology* Physics},
  volume={73},
  number={5},
  pages={1364--1368},
  year={2009},
  publisher={Elsevier}
}

@article{wagner2000geometrically,
  title={A geometrically based method for automated radiosurgery planning},
  author={Wagner, Thomas H and Yi, Taeil and Meeks, Sanford L and Bova, Francis J and Brechner, Beverly L and Chen, Yunmei and Buatti, John M and Friedman, William A and Foote, Kelly D and Bouchet, Lionel G},
  journal={International Journal of Radiation Oncology* Biology* Physics},
  volume={48},
  number={5},
  pages={1599--1611},
  year={2000},
  publisher={Elsevier}
}

@article{alizadeh2020electromyography,
  title={An electromyography-assisted biomechanical cervical spine model: model development and validation},
  author={Alizadeh, Mina and Aurand, Alexander and Knapik, Gregory G and Dufour, Jonathan S and Mendel, Ehud and Bourekas, Eric and Marras, William S},
  journal={Clinical Biomechanics},
  volume={80},
  pages={105169},
  year={2020},
  publisher={Elsevier}
}

@article{bogduk2000biomechanics,
  title={Biomechanics of the cervical spine. I: Normal kinematics},
  author={Bogduk, Nikolai and Mercer, Susan},
  journal={Clinical biomechanics},
  volume={15},
  number={9},
  pages={633--648},
  year={2000},
  publisher={Elsevier}
}

@article{wiersma2013spatial,
  title={Spatial and temporal performance of 3D optical surface imaging for real-time head position tracking},
  author={Wiersma, Rodney D and Tomarken, SL and Grelewicz, Zachary and Belcher, Andrew H and Kang, Hyejoo},
  journal={Medical physics},
  volume={40},
  number={11},
  pages={111712},
  year={2013},
  publisher={Wiley Online Library}
}

@article{belcher2014development,
  title={Development of a 6DOF robotic motion phantom for radiation therapy},
  author={Belcher, Andrew H and Liu, Xinmin and Grelewicz, Zachary and Pearson, Erik and Wiersma, Rodney D},
  journal={Medical physics},
  volume={41},
  number={12},
  pages={121704},
  year={2014},
  publisher={Wiley Online Library}
}

@article{suderman2012neck,
  title={Neck muscle paths and moment arms are significantly affected by wrapping surface parameters},
  author={Suderman, Bethany L and Krishnamoorthy, Bala and Vasavada, Anita N},
  journal={Computer methods in biomechanics and biomedical engineering},
  volume={15},
  number={7},
  pages={735--744},
  year={2012},
  publisher={Taylor \& Francis}
}

@article{zhang2019robotic,
  title={A robotic neck brace to characterize head-neck motion and muscle electromyography in subjects with amyotrophic lateral sclerosis},
  author={Zhang, Haohan and Chang, Biing-Chwen and Andrews, Jinsy and Mitsumoto, Hiroshi and Agrawal, Sunil},
  journal={Annals of clinical and translational neurology},
  volume={6},
  number={9},
  pages={1671--1680},
  year={2019},
  publisher={Wiley Online Library}
}

@article{zaitsev2015motion,
  title={Motion artifacts in MRI: A complex problem with many partial solutions},
  author={Zaitsev, Maxim and Maclaren, Julian and Herbst, Michael},
  journal={Journal of Magnetic Resonance Imaging},
  volume={42},
  number={4},
  pages={887--901},
  year={2015},
  publisher={Wiley Online Library}
}

@article{gurney2018mri,
  title={MRI-based assessment of 3D Intrafractional motion of head and neck cancer for radiation therapy},
  author={Gurney-Champion, Oliver J and McQuaid, Dualta and Dunlop, Alex and Wong, Kee H and Welsh, Liam C and Riddell, Angela M and Koh, Dow-Mu and Oelfke, Uwe and Leach, Martin O and Nutting, Christopher M and others},
  journal={International Journal of Radiation Oncology* Biology* Physics},
  volume={100},
  number={2},
  pages={306--316},
  year={2018},
  publisher={Elsevier}
}

@article{liu2018effect,
  title={Effect of head motion on MRI B 0 field distribution},
  author={Liu, Jiaen and de Zwart, Jacco A and van Gelderen, Peter and Murphy-Boesch, Joseph and Duyn, Jeff H},
  journal={Magnetic resonance in medicine},
  volume={80},
  number={6},
  pages={2538--2548},
  year={2018},
  publisher={Wiley Online Library}
}

@article{konz2006kinematic,
  title={A kinematic model to assess spinal motion during walking},
  author={Konz, Regina J and Fatone, Stefania and Stine, Rebecca L and Ganju, Aruna and Gard, Steven A and Ondra, Stephen L},
  journal={Spine},
  volume={31},
  number={24},
  pages={E898--E906},
  year={2006},
  publisher={LWW}
}

@article{spin2016patient,
  title={Patient movement and motion artefacts in cone beam computed tomography of the dentomaxillofacial region: a systematic literature review},
  author={Spin-Neto, Rubens and Wenzel, Ann},
  journal={Oral surgery, oral medicine, oral pathology and oral radiology},
  volume={121},
  number={4},
  pages={425--433},
  year={2016},
  publisher={Elsevier}
}

@article{stutzki2015indications,
  title={Indications of cone beam CT in head and neck imaging},
  author={Stutzki, Malene and Jahns, Evelyn and Mandapathil, Magis Matthew and Diogo, Isabell and Werner, Jochen Alfred and G{\"u}ldner, Christian},
  journal={Acta oto-laryngologica},
  volume={135},
  number={12},
  pages={1337--1343},
  year={2015},
  publisher={Taylor \& Francis}
}

@article{krist2022management,
  title={Management of brain metastasis. Surgical resection versus stereotactic radiotherapy: A meta-analysis},
  author={Krist, David T and Naik, Anant and Thompson, Charee M and Kwok, Susanna S and Janbahan, Mika and Olivero, William C and Hassaneen, Wael},
  journal={Neuro-Oncology Advances},
  volume={4},
  number={1},
  pages={vdac033},
  year={2022},
  publisher={Oxford University Press US}
}

@article{gaspar2010role,
  title={The role of whole brain radiation therapy in the management of newly diagnosed brain metastases: a systematic review and evidence-based clinical practice guideline},
  author={Gaspar, Laurie E and Mehta, Minesh P and Patchell, Roy A and Burri, Stuart H and Robinson, Paula D and Morris, Rachel E and Ammirati, Mario and Andrews, David W and Asher, Anthony L and Cobbs, Charles S and others},
  journal={Journal of neuro-oncology},
  volume={96},
  pages={17--32},
  year={2010},
  publisher={Springer}
}

@article{graber2019congress,
  title={Congress of neurological surgeons systematic review and evidence-based guidelines on the use of stereotactic radiosurgery in the treatment of adults with metastatic brain tumors},
  author={Graber, Jerome J and Cobbs, Charles S and Olson, Jeffrey J},
  journal={Neurosurgery},
  volume={84},
  number={3},
  pages={E168--E170},
  year={2019},
  publisher={LWW}
}

@article{chelvarajah2004cranial,
  title={Cranial immobilisation—is there a better way?},
  author={Chelvarajah, Radhini and Leighton, Brigid and Martin, Linda and Smith, Wayne and Beldham-Collins, Rachael},
  journal={Radiographer},
  volume={51},
  number={1},
  pages={29--33},
  year={2004},
  publisher={Wiley Online Library}
}

@article{godenschweger2016motion,
  title={Motion correction in MRI of the brain},
  author={Godenschweger, Frank and K{\"a}gebein, Urte and Stucht, Daniel and Yarach, Uten and Sciarra, Alessandro and Yakupov, Renat and L{\"u}sebrink, Falk and Schulze, Peter and Speck, Oliver},
  journal={Physics in medicine \& biology},
  volume={61},
  number={5},
  pages={R32},
  year={2016},
  publisher={IOP Publishing}
}

@article{zahneisen2016reverse,
  title={Reverse retrospective motion correction},
  author={Zahneisen, Benjamin and Keating, Brian and Singh, Aditya and Herbst, Michael and Ernst, Thomas},
  journal={Magnetic resonance in medicine},
  volume={75},
  number={6},
  pages={2341--2349},
  year={2016},
  publisher={Wiley Online Library}
}

@article{banerjee2013parallel,
  title={Parallel and partial Fourier imaging with prospective motion correction},
  author={Banerjee, Suchandrima and Beatty, Philip J and Zhang, Jian Z and Shankaranarayanan, Ajit},
  journal={Magnetic Resonance in Medicine},
  volume={69},
  number={2},
  pages={421--433},
  year={2013},
  publisher={Wiley Online Library}
}

@article{williams2023ultra,
  title={Ultra-high field MRI: parallel-transmit arrays and RF pulse design},
  author={Williams, Sydney N and McElhinney, Paul and Gunamony, Shajan},
  journal={Physics in Medicine \& Biology},
  volume={68},
  number={2},
  pages={02TR02},
  year={2023},
  publisher={IOP Publishing}
}

@article{wang2023diffusion,
  title={Diffusion-Weighted Imaging of the Abdomen: Correction for Gradient Nonlinearity Bias in Apparent Diffusion Coefficient},
  author={Wang, Jian and Ma, Chao and Yang, Panpan and Wang, Zhen and Chen, Yufei and Bian, Yun and Shao, Chengwei and Lu, Jianping},
  journal={Journal of Magnetic Resonance Imaging},
  volume={58},
  number={1},
  pages={223--231},
  year={2023},
  publisher={Wiley Online Library}
}

@article{he2023frameless,
  title={Frameless versus frame-based stereotactic radiosurgery for intracranial arteriovenous malformations: a propensity-matched analysis},
  author={He, Zhexi and Ho, Man Kit Jason and Lee, Wan Yan Venus and Law, Hing Yuen and Wong, Yee Wa Victy and Leung, To-wai and Mui, Wing Ho and Wong, Sui-To and Wong, Chi Sing Frank and Yam, Kwong Yui},
  journal={Clinical and Translational Radiation Oncology},
  volume={41},
  pages={100642},
  year={2023},
  publisher={Elsevier}
}

@article{almeida2020positioning,
  title={Positioning deviations in frameless and frame-based intracranial stereotactic radiosurgery},
  author={Almeida, TVR and Da Silva, CM and Piedade, PA and Cordova Jr, AL and Bezerra Jr, CG and Soboll, DS and Denyak, V and Schelin, HR},
  journal={Radiation Physics and Chemistry},
  volume={167},
  pages={108363},
  year={2020},
  publisher={Elsevier}
}

@article{takakura2009geometric,
  title={The geometric accuracy of frameless stereotactic radiosurgery using a 6D robotic couch system},
  author={Takakura, T and Mizowaki, T and Nakata, M and Yano, S and Fujimoto, T and Miyabe, Y and Nakamura, M and Hiraoka, M},
  journal={Physics in Medicine \& Biology},
  volume={55},
  number={1},
  pages={1},
  year={2009},
  publisher={IOP Publishing}
}

@article{li2013migration,
  title={Migration from full-head mask to “open-face” mask for immobilization of patients with head and neck cancer},
  author={Li, Guang and Lovelock, D Michael and Mechalakos, James and Rao, Shyam and Della-Biancia, Cesar and Amols, Howard and Lee, Nancy},
  journal={Journal of Applied Clinical Medical Physics},
  volume={14},
  number={5},
  pages={243--254},
  year={2013},
  publisher={Wiley Online Library}
}

@article{copeland2021infant,
  title={Infant and child MRI: a review of scanning procedures},
  author={Copeland, Anni and Silver, Eero and Korja, Riikka and Lehtola, Satu J and Merisaari, Harri and Saukko, Ekaterina and Sinisalo, Susanne and Saunavaara, Jani and L{\"a}hdesm{\"a}ki, Tuire and Parkkola, Riitta and others},
  journal={Frontiers in neuroscience},
  volume={15},
  pages={666020},
  year={2021},
  publisher={Frontiers Media SA}
}

@article{woods2010training,
  title={Training children to reduce motion and increase success of MRI scanning},
  author={Woods-Frohlich, Lindsay and Martin, Toby and L. Malisza, Krisztina},
  journal={Current Medical Imaging Reviews},
  volume={6},
  number={3},
  pages={165--170},
  year={2010},
  publisher={Bentham Science Publishers direct}
}

@article{li2023pet,
  title={PET/CT scan without sedation: how to use total-body PET/CT to salvage child’s involuntary movement?},
  author={Li, Yongjiang and Wang, Juan and Hu, Junfeng and Jia, Jin and Sun, Hongyan and Zhao, Yumo and Hu, Yingying},
  journal={European Journal of Nuclear Medicine and Molecular Imaging},
  volume={50},
  number={9},
  pages={2912--2913},
  year={2023},
  publisher={Springer}
}

@article{xue2017high,
  title={High-accuracy and real-time 3D positioning, tracking system for medical imaging applications based on 3D digital image correlation},
  author={Xue, Yuan and Cheng, Teng and Xu, Xiaohai and Gao, Zeren and Li, Qianqian and Liu, Xiaojing and Wang, Xing and Song, Rui and Ju, Xiangyang and Zhang, Qingchuan},
  journal={Optics and Lasers in Engineering},
  volume={88},
  pages={82--90},
  year={2017},
  publisher={Elsevier}
}

@article{dellios2021evaluation,
  title={Evaluation of patient-specific MR distortion correction schemes for improved target localization accuracy in SRS},
  author={Dellios, Dimitrios and Pappas, Eleftherios P and Seimenis, Ioannis and Paraskevopoulou, Chryssa and Lampropoulos, Kostas I and Lymperopoulou, Georgia and Karaiskos, Pantelis},
  journal={Medical Physics},
  volume={48},
  number={4},
  pages={1661--1672},
  year={2021},
  publisher={Wiley Online Library}
}

@article{vergalasova2019multi,
  title={Multi-institutional dosimetric evaluation of modern day stereotactic radiosurgery (SRS) treatment options for multiple brain metastases},
  author={Vergalasova, Irina and Liu, Haisong and Alonso-Basanta, Michelle and Dong, Lei and Li, Jun and Nie, KE and Shi, Wenyin and Teo, Boon-Keng Kevin and Yu, Yan and Yue, Ning Jeff and others},
  journal={Frontiers in oncology},
  volume={9},
  pages={483},
  year={2019},
  publisher={Frontiers Media SA}
}

@article{yoganandan2001biomechanics,
  title={Biomechanics of the cervical spine Part 2. Cervical spine soft tissue responses and biomechanical modeling},
  author={Yoganandan, Narayan and Kumaresan, Srirangam and Pintar, Frank A},
  journal={Clinical biomechanics},
  volume={16},
  number={1},
  pages={1--27},
  year={2001},
  publisher={Elsevier}
}

@article{swartz2005cervical,
  title={Cervical spine functional anatomy and the biomechanics of injury due to compressive loading},
  author={Swartz, Erik E and Floyd, RT and Cendoma, Mike},
  journal={Journal of athletic training},
  volume={40},
  number={3},
  pages={155},
  year={2005}
}

@article{keeling2016quantitative,
  title={Quantitative evaluation of patient setup uncertainty of stereotactic radiotherapy with the frameless 6D ExacTrac system using statistical modeling},
  author={Keeling, Vance and Hossain, Sabbir and Jin, Hosang and Algan, Ozer and Ahmad, Salahuddin and Ali, Imad},
  journal={Journal of Applied Clinical Medical Physics},
  volume={17},
  number={3},
  pages={111--127},
  year={2016},
  publisher={Wiley Online Library}
}
\end{document}